\documentclass{article}

\usepackage[preprint]{neurips_2026}

\usepackage[utf8]{inputenc} % allow utf-8 input
\usepackage[T1]{fontenc}    % use 8-bit T1 fonts
\usepackage{hyperref}       % hyperlinks
\usepackage{url}            % simple URL typesetting
\usepackage{booktabs}       % professional-quality tables
\usepackage{amsfonts}       % blackboard math symbols
\usepackage{amsmath}
\usepackage{amssymb}
\usepackage{nicefrac}       % compact symbols for 1/2, etc.
\usepackage{microtype}      % microtypography
\usepackage{xcolor}         % colors
\usepackage{graphicx}       % figures
\usepackage{enumitem}       % nicer lists
\usepackage{subcaption}     % subfigures (sub-panel labels)
\usepackage{multirow}       % multi-row table cells

\title{When RL Suppresses Its Own Vocabulary:\\
Recovering Reasoning Diversity in Puzzle-to-Math Transfer}

\author{%
  Mayug Maniparambil$^{1}$\thanks{Correspondence: mayugmaniparambil@gmail.com} \And
  Arjun Karuvally$^{2}$ \And
  Terrence Sejnowski$^{2}$ \And
  Fergal Reid$^{1}$ \\[0.5em]
  $^{1}$Fin AI Research \quad $^{2}$Salk Institute for Biological Studies
}

\begin{document}

\maketitle

\begin{abstract}
Reinforcement learning using verifiable rewards (RLVR) improves LLM reasoning, but the conditions
under which it transfers across domains---and why it does so---remain under-explored. We study
cross-domain transfer in a 7B model whose SFT and RL post-training stages use
only constraint-satisfaction puzzles, with no mathematics problems in the
post-training data. To analyze how transfer emerges, we introduce a reasoning primitive-level framework that combines a 9-class span classifier with motif extraction, allowing us to segment chain-of-thought traces into primitive motifs and track their evolution across training stages and domains. We find that puzzle SFT induces a reasoning-primitive vocabulary, yielding a $+7$pp \texttt{pass@32} gain on OlymMATH-Hard. Vanilla GSPO then composes these primitives into longer compute--verify chains, adding a further $+6$pp. However, this RL stage also suppresses exploratory primitives such as \textit{hypothesize} and \textit{backtrack}. To address this, we introduce a novelty bonus that rewards diverse correct rollouts, using perplexity under the reference model as a signal. This restores recovery primitives during RL and adds a further $+7$pp
\texttt{pass@32} relative to vanilla GSPO. Finally, the end-to-end recipe raises the hard-math capability ceiling from $16.0\%$ at the OLMo3-7B-Instruct-SFT base to $36.0\%$, without adding any
mathematics problems during the SFT or RL stages.
\end{abstract}

% Section files in sections/ (one per logical section). Reorder, swap, or replace
% individual sections by editing the corresponding sections/NN_*.tex file.
% Section 1: Introduction
% Drafted 2026-05-02 using the writing/neurips_paper/intro_guide.txt template.
% Structure: problem framing + intuition / gap / idea / evidence / scope / contributions.

\section{Introduction}
\label{sec:intro}

Reinforcement learning has driven much of the recent progress in large language model (LLM) reasoning on math and code, but how much of this improvement reflects general problem-solving capability versus skills specific to the training distribution is unclear. 
We probe this in a setting designed to dissociate the two: a 7B model is trained with supervised fine-tuning (SFT) and reinforcement learning (RL) on constraint-satisfaction puzzles only — no mathematics problems appear in post-training — and is evaluated on hard mathematics benchmarks where current models remain far from saturation. 
Puzzles and mathematics share problem-solving operations such as search, planning, verification, and backtracking \citep{polya1945how,simon1971human,schoenfeld1985mathematical}, while differing substantially in problem framing and notation. 
Whatever RL contributes to math performance in this setup cannot come from post-training exposure to mathematics. If transfer occurs, it must reflect changes to the abstract operations the two domains share. This setup lets us ask: does RL on puzzles improve hard-mathematics performance, and if so, what reasoning-level changes are responsible?

Transfer to math occurs: after SFT and RL on puzzles, \texttt{pass@32} on OlymMATH-Hard rises by $20$ percentage points (from $16.0\%$ to $36.0\%$) without any mathematics in post-training. 
To understand what reasoning-level changes drive this gain, we develop a primitive-level analysis framework. 
A span classifier segments chain-of-thought reasoning into 9 primitives --- \textsc{hypothesize}, \textsc{compute}, \textsc{check}, \textsc{backtrack}, \textsc{plan}, \textsc{enumerate}, \textsc{setup}, \textsc{summarize}, \textsc{other} --- and a motif-extraction step characterizes how these primitives compose into recurring sequences (motifs). 
The framework reveals a depth-recovery tradeoff: RL extends \textit{depth} (longer \textsc{compute}–\textsc{check} motif chains) but suppresses \textit{recovery} (\textsc{hypothesize} and \textsc{backtrack} drop by $70$--$80\%$).

This depth-recovery tradeoff we identified motivates a modification to the RL reward: a perplexity-based novelty bonus that rewards rollouts unlikely under a frozen reference model, designed to preserve the recovery primitives vanilla RL suppresses. The bonus restores \textsc{hypothesize} and \textsc{backtrack} while retaining the depth gains from the \textsc{compute}–\textsc{verify} chains, adding a further 7pp \texttt{pass@32} over a matched vanilla baseline. On problems that only the novelty-trained model solves, recovery primitives appear at substantially higher rates, suggesting the bonus works by keeping the model flexible enough to change course when direct execution fails.

% There have been notable recent efforts on puzzle-to-math transfer that have left the central question we ask in the paper unanswered.
Our conclusions --- that RL on puzzles transfers, that it does so via primitive-level changes, and that a novelty bonus amplifies the effect --- rely on a setup that cleanly isolates RL's contribution. Recent work on puzzle-to-math transfer has not had this property. 
\textbf{Enigmata}~\citep{chen2025enigmata} trains on a 36-puzzle
suite at scale, but includes math problems in
both SFT and RL, making it difficult to isolate whether transfer performance originates from in-domain math
exposure. \textbf{Logic-RL}~\citep{xie2025logicrl} and
\textbf{LogicPuzzleRL}~\citep{wong2025logicpuzzlerl} train on puzzles only but
evaluate primarily at \texttt{pass@1} on benchmarks no harder than AIME24.
At \texttt{pass@1}, it is impossible to distinguish whether RL expands the set of
problems the model can solve at all --- capability ceiling expansion --- or merely
sharpens a solution path it could already reach under sufficient sampling.
Whether puzzle-only RL can raise the hard-math capability ceiling therefore
remains undemonstrated in the prior literature.

\paragraph{Contributions}: (1) We show that RL on a non-mathematical domain (constraint-satisfaction puzzles) transfers to hard mathematics benchmarks, isolating RL's contribution from SFT, (2) We introduce a primitive- and motif-level analysis framework for chain-of-thought reasoning, and use it to identify a depth–recovery tradeoff induced by vanilla RL, (3) We propose a perplexity-based novelty bonus that resolves this tradeoff and yields a further 7pp \texttt{pass@32}.

% Main pass@k figure placed after the intro so it can float to page 2.
\begin{figure}[t!]
  \centering
  \begin{subfigure}{0.32\linewidth}
    \centering
    \includegraphics[width=\linewidth]{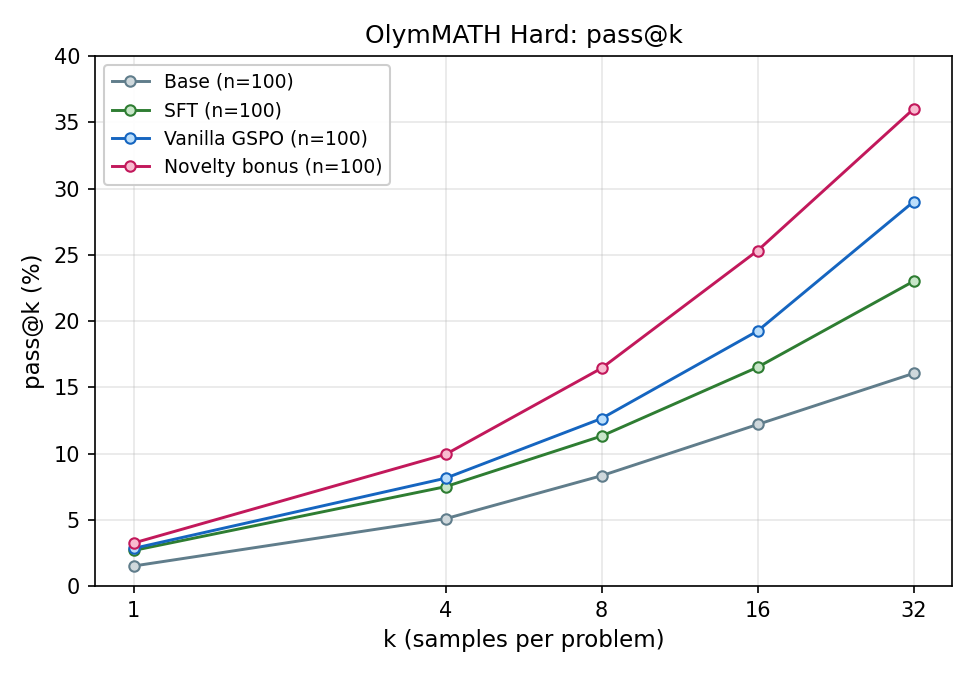}
    \caption{OlymMATH-Hard (primary)}
    \label{fig:main-passk-hard}
  \end{subfigure}\hfill
  \begin{subfigure}{0.32\linewidth}
    \centering
    \includegraphics[width=\linewidth]{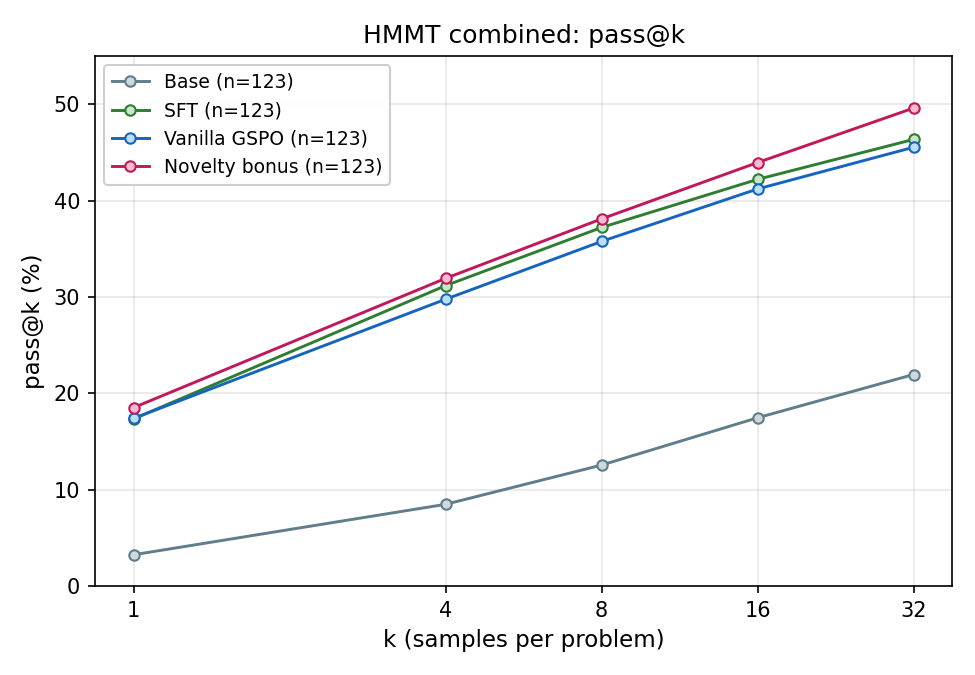}
    \caption{HMMT combined ($N{=}123$)}
    \label{fig:math-passk-hmmt}
  \end{subfigure}\hfill
  \begin{subfigure}{0.32\linewidth}
    \centering
    \includegraphics[width=\linewidth]{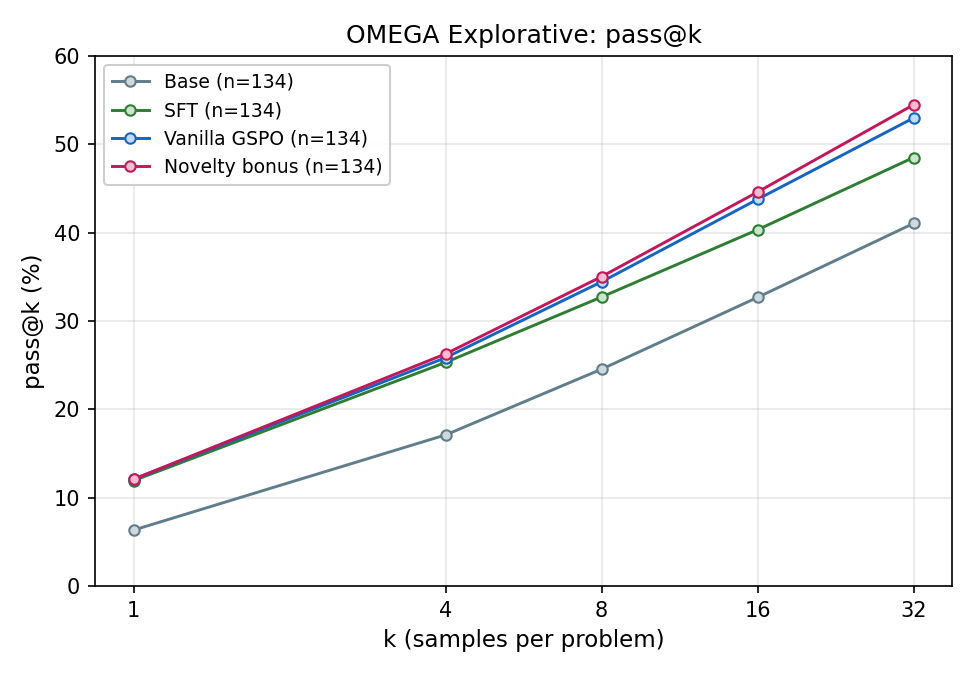}
    \caption{OMEGA Explorative ($N{=}134$)}
    \label{fig:math-passk-omega}
  \end{subfigure}\\[4pt]
  \begin{subfigure}{0.32\linewidth}
    \centering
    \includegraphics[width=\linewidth]{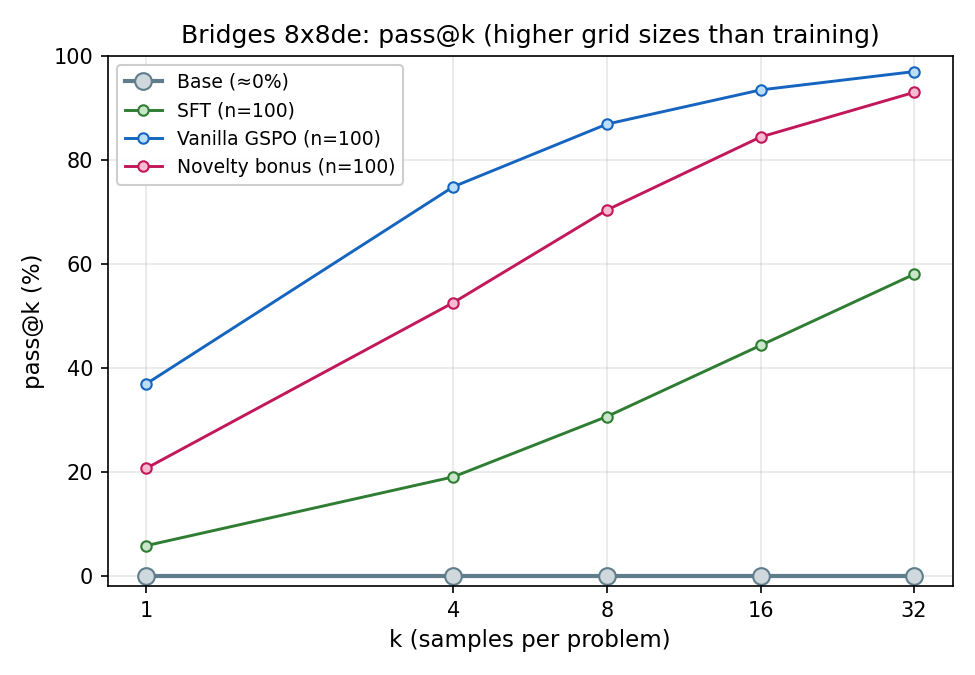}
    \caption{Bridges $8{\times}8$}
    \label{fig:main-passk-bridges}
  \end{subfigure}\hfill
  \begin{subfigure}{0.32\linewidth}
    \centering
    \includegraphics[width=\linewidth]{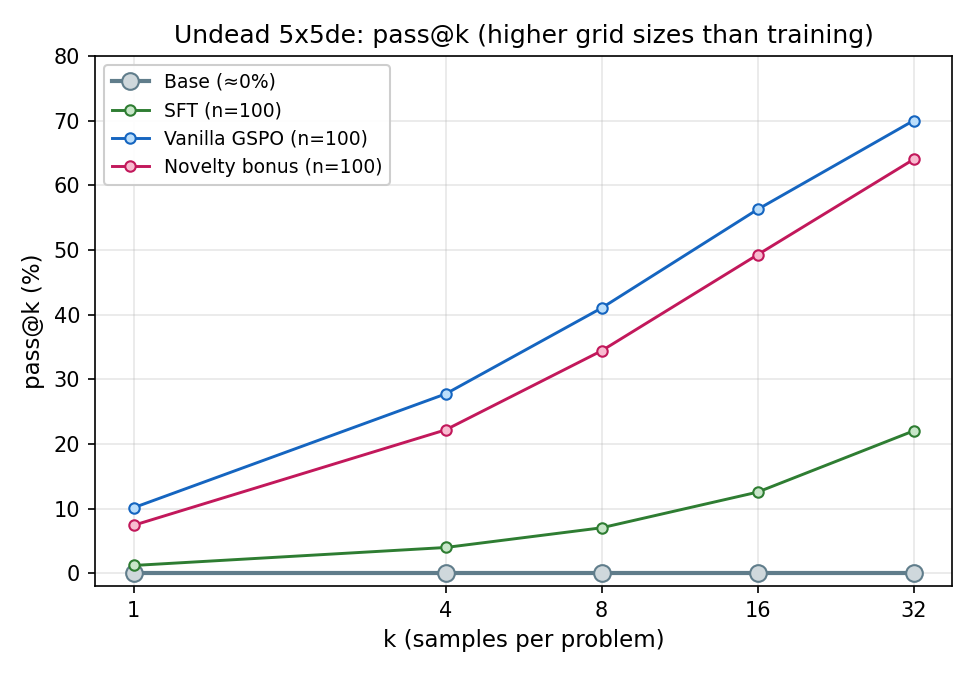}
    \caption{Undead $5{\times}5$}
    \label{fig:main-passk-undead}
  \end{subfigure}\hfill
  \begin{subfigure}{0.32\linewidth}
    \centering
    \includegraphics[width=\linewidth]{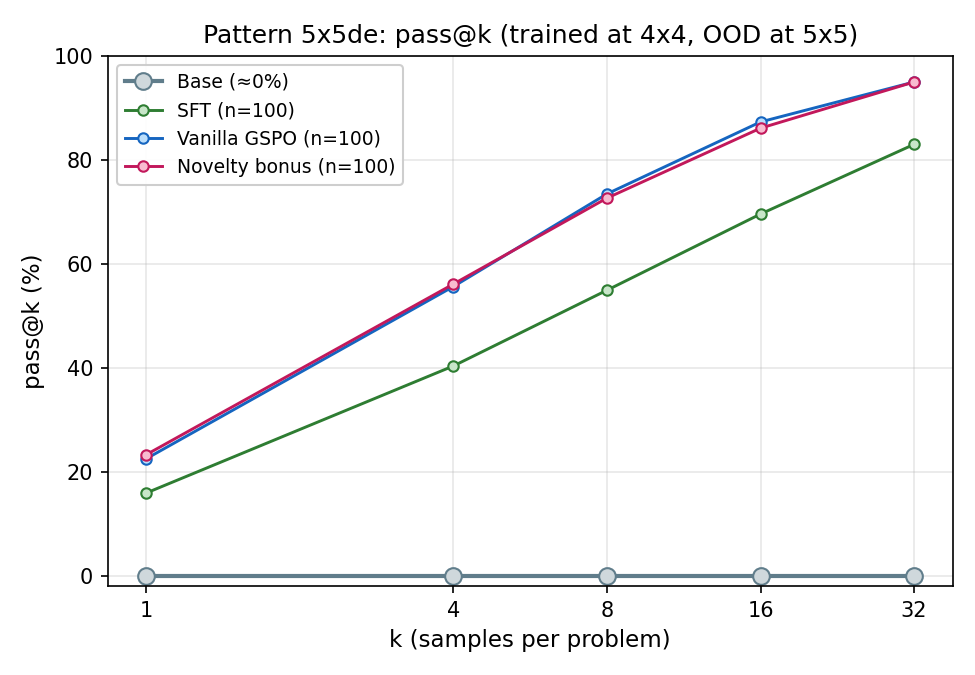}
    \caption{Pattern $5{\times}5$}
    \label{fig:puzzle-passk-pattern}
  \end{subfigure}
  \caption{\texttt{pass@}$k$ across the recipe. Top: held-out math benchmarks;
    bottom: larger held-out puzzle grids than those used in training.
    Puzzle-only post-training raises \texttt{pass@}$k$ on both math and
    higher-grid-size puzzles; additional math benchmarks are in
    Appendix~\ref{app:additional-benchmarks}.}
  \label{fig:math-passk}
\end{figure}

% Section 2: Related Work
%
% Three subsections (locked 2026-05-01):
%   2.1  RL for LLM reasoning, incl. synthetic/puzzle tasks  -- DRAFTED
%   2.2  Exploration and diversity in RL fine-tuning         -- TODO
%   2.3  Mechanism analyses of reasoning                     -- TODO
%
% Lit search files: writing/lit_search_{rl_puzzles,diversity,mechanism}.md
% Bib keys below are placeholders; entries to be added to references.bib.

\section{Related Work}
\label{sec:related}

\paragraph{RL for LLM reasoning and synthetic puzzle tasks.}
Reinforcement learning from verifiable rewards (RLVR), popularised by
DeepSeek-R1~\cite{guo2025deepseekr1} and developed through objectives such as
GRPO~\cite{shao2024deepseekmath}, GSPO~\cite{zheng2025gspo}, and
DAPO~\cite{yu2025dapo}, has become a standard recipe for post-training
reasoning models. Synthetic puzzles with deterministic verifiers are a natural
testbed for this recipe, and recent works such as Enigmata~\cite{chen2025enigmata},
Logic-RL~\cite{xie2025logicrl}, and
LogicPuzzleRL~\cite{wong2025logicpuzzlerl} report transfer from puzzle training
to mathematical reasoning. However, existing evidence either mixes math data
into the training recipe or evaluates puzzle-only training primarily at
\texttt{pass@1} on less difficult math benchmarks, leaving open whether
non-math RL can expand the mathematical reasoning ceiling, measured by
\texttt{pass@k} at large $k$. This question connects to a broader debate about
RL post-training transfer: some work reports broad cross-domain gains from math
RL~\cite{chengrevisiting}, while other studies find that single-domain RL mainly
sharpens in-domain behaviour~\cite{huang2025mathreasoning,breakingbarriers2025,yue2025limitrlvr}.
We study this in a clean isolation: our pipeline uses no math data in SFT or RL
and evaluates \texttt{pass@k} on hard out-of-domain math.

\paragraph{Exploration and diversity in RL fine-tuning.}
Recent work has argued that RLVR can improve \texttt{pass@1} while reducing
\texttt{pass@k} diversity, motivating method families that preserve exploration during
RL fine-tuning~\cite{yue2025limitrlvr,cui2025entropy,passkdiagnostic2025, chen2025pass}.
One family modifies token-level signals via covariance-based clipping, entropy-term additions, or low-probability regularisation~\cite{cui2025entropy,cheng2025reasoningexploration,huang2025lpreg}. A second
family reweights whole trajectories by rarity, using current-policy ranks,
strategy clusters, or outcome-level novelty
~\cite{he2025unlikeliness,hu2026rare,song2025obe}. Our novelty bonus belongs to the trajectory-level family but differs from prior
methods in its anchor: novelty is scored under a frozen reference model rather
than the current policy, so the diversity target does not drift as training
progresses. The bonus applies only to verifier-correct rollouts and uses top-$k$
token NLL rather than LLM-judge clustering, rewarding alternative successful
reasoning paths without per-trace judge calls at training time. Unlike work that optimises
\texttt{pass@k} directly~\cite{walder2025pass}, we retain the standard
verifiable-reward objective and use novelty as an auxiliary
exploration-preserving bonus.

\paragraph{Mechanism analyses of reasoning.}
A growing literature analyzes reasoning mechanisms by labeling behaviors in
chain-of-thought traces, including verification, backtracking, subgoal setting,
and related cognitive operations~\cite{gandhi2025fourhabits,kim2025astro}.
These studies show that RL reshapes the distribution of reasoning primitives,
amplifying some and suppressing others~\cite{gandhi2025fourhabits,howwhygeneralize2025}.
Our analysis differs in scale and method: we distill judge annotations into a
lightweight span classifier that can label $25$k-token traces across tens of
thousands of rollouts and ten checkpoints, avoiding both per-trace judge calls
and injected behavior markers. Unlike prior analyses, we use the resulting motif
distributions to study out-of-domain transfer rather than tracking only
per-trace primitive counts.

% Section 3: Setup, Recipe, and Main Results (compact rewrite)
% Original verbose version preserved in sections/03_setup_recipe_results.tex.

\section{Post-Training Pipeline and Results}
\label{sec:results}

\paragraph{Setup and evaluation.}
We use OLMo3-7B-Instruct-SFT~\cite{olmo2025olmo3} as our base model: it is
fully open, has undergone no prior RL fine-tuning, and retains 
upstream mathematical knowledge from pretraining and instruction tuning. This last property is
critical as any post-training gain reflects improved search and verification
over existing knowledge, not the injection of new mathematical content.
Our post-training pipeline uses only constraint-satisfaction puzzles from the Simon Tatham
collection~\cite{tatham_puzzles,estermann2024puzzles,maniparambil2026topobench} --- Bridges, Pattern,
Undead, and Galaxies --- with no mathematics problems in either the SFT or RL
stages. We evaluate on three hard out-of-domain mathematics benchmarks:
OlymMATH-Hard~\cite{sun2025olympmath}, HMMT~\cite{balunovic2025matharena}, and
OMEGA~\cite{sun2025omega}, all of which remain far from saturation for a 7B model and therefore probe the capability ceiling rather than benchmark
familiarity. Our primary metric is \texttt{pass@32}, since a \texttt{pass@1} improvement can
reflect either capability ceiling expansion, where new problems become solvable,
or sharpening of an already-reachable path, and the two are indistinguishable at
the single-sample level. Large-$k$ pass rates isolate the former, and all
evaluations use $32$ rollouts per problem at temperature $T{=}0.6$.

\paragraph{Training Recipe.}
The training pipeline has two stages. First, because the base model has near-zero puzzle
competence, direct verifier RL would provide little reward variance, so we
perform supervised fine-tuning on rejection-sampled reasoning traces distilled
from Deep-Seek Reasoner (DSR), keeping only traces that verify as correct under the puzzle scorers.
Second, from this SFT checkpoint we train two multi-puzzle
variants of the RL algorithm Group Sequence Policy Optimization (GSPO)~\cite{zheng2025gspo}. 
Both variants use the same verifier reward, which
combines an exact reward for fully correct boards with a partial completeness
reward. The partial completeness reward is proportional to how much of the final board state matches the solution.
The vanilla GSPO condition uses only this verifier reward, while the novelty
condition additionally rewards correct rollouts whose reasoning traces are
assigned low probability under the frozen SFT reference. We use GSPO rather
than token-level Group Relative Policy Optimization (GRPO) because our target responses are $25$--$28$k tokens and
GSPO trained reliably in long-context pilots where GRPO was unstable. Full
hyperparameters and reward details are in Appendix~\ref{app:hyperparams}.

\paragraph{Results.}
Puzzle SFT alone raises OlymMATH-Hard \texttt{pass@32} from $16.0\%$ at the
base to $23.0\%$. Both RL variants, trained from this SFT checkpoint, improve
further: vanilla GSPO reaches $29.0\%$ ($+6$pp over SFT), while the novelty
variant reaches $\mathbf{36.0\%}$ ($+13$pp over SFT and $+7$pp over the matched
vanilla GSPO baseline), as shown in Figure~\ref{fig:math-passk} and
Table~\ref{tab:main}. Notably, the novelty gain grows with $k$---modest at
\texttt{pass@1} but $+7$pp at \texttt{pass@32}---showing that it
increases the capability ceiling rather than
sharpening a dominant path. The same pattern holds across other tested benchmarks: on
OMEGA Explorative~\cite{sun2025omega}, a synthetic benchmark of structured
mathematical problems at increased complexity, the novelty variant reaches
$54.5\%$ against $48.5\%$ for SFT; on HMMT~\cite{balunovic2025matharena} ($123$
competition problems), it reaches $49.6\%$ against $46.3\%$. Per-problem solve rates further support this reading (Appendix~Figure~\ref{fig:per-problem-progression}): Base$\to$Novelty newly solves $29$
of $100$ OlymMATH-Hard problems while regressing on only $5$, confirming the
recipe lifts the ceiling broadly rather than sharpening a few already-reachable
problems. Finally, on larger-grid puzzle evaluations, both RL variants raise
\texttt{pass@}$k$ over SFT across the full sampling budget on Bridges
$8{\times}8$ and Undead $5{\times}5$, suggesting the learned strategies
extrapolate across grid scale.

The remainder of the paper asks why puzzle-only training transfers to hard
mathematics: what reasoning behaviors does SFT introduce, what does
vanilla RL amplify or suppress, and which changes predict the sampled ceiling
gains observed above? Section~\ref{sec:framework} introduces the primitive-level
analysis, Sections~\ref{sec:sft-induces}--\ref{sec:rl-collapses} apply it to
the recipe stages, and Sections~\ref{sec:novelty}--\ref{sec:confirmation} use
the resulting diagnosis to derive and validate the novelty bonus.

% Section 4: Analysis Framework (supporting tool, not primary contribution)

\section{Analysis Framework: Reasoning Primitives}
\label{sec:framework}

Aggregate metrics such as response length and token entropy reveal that training
changes the policy~\cite{shypula2025evaluating,deshpande2025diverse}, but not
\emph{what} reasoning behaviours change. We therefore analyse chain-of-thought
traces as sequences of reasoning primitives: short, semantically coherent spans
that play canonical roles such as planning, computing, checking, hypothesising,
or backtracking. Because traces can reach $\sim\!25$k tokens, we split each
trace into spans at paragraph boundaries and discourse markers (e.g.,
\textit{Case 1}, \textit{Wait}, \textit{Alternatively}, \textit{Going back}),
yielding contiguous spans of $80$--$250$ tokens (full algorithm in
Appendix~\ref{app:motif-details}). Each span is then classified with its
preceding context, producing one primitive sequence per rollout.

\begin{table}[t]
  \caption{Primitive vocabulary used by the span classifier.}
  \label{tab:primitives}
  \centering
  \footnotesize
  \setlength{\tabcolsep}{4pt}
  \begin{tabular}{lp{0.26\linewidth}lp{0.26\linewidth}}
    \toprule
    Primitive & Meaning & Primitive & Meaning \\
    \midrule
    \textsc{plan}        & High-level approach. &
    \textsc{setup}       & Notation or auxiliaries. \\
    \textsc{enumerate}   & Cases or candidates. &
    \textsc{hypothesize} & Tentative claim to test. \\
    \textsc{compute}     & Calculation or manipulation. &
    \textsc{check}       & Verify against constraints. \\
    \textsc{backtrack}   & Reject and return. &
    \textsc{summarize}   & Consolidate partial results. \\
    \textsc{other}       & Filler or formatting. &
                         & \\
    \bottomrule
  \end{tabular}
\end{table}

Table~\ref{tab:primitives} gives the nine-class vocabulary. We developed the
labels using a jury of LLM judges (DeepSeek-V3.2 and
DeepSeek-Chat~\cite{liu2025deepseekv32}), iterating until inter-judge agreement
reached $86\%$ on a held-out span set. Because judge-based labelling is
impractical at scale, we distil the judges into a nine-way span classifier that
reaches $0.80$ macro-F1 against judge consensus.

Finally, we convert each primitive sequence into motifs --- $k$-grams of
primitives for $k = 2, \ldots, 15$ --- and for each checkpoint track primitive
frequencies, motif frequencies, and their cross-checkpoint correlation with
\texttt{pass@}$k$ (Appendix~\ref{app:motif-details}). Tracking these statistics
across checkpoints lets us identify which reasoning patterns each training stage
amplifies or suppresses, and which correlate with hard-math ceiling gains.

% Section 5: SFT Induces Reasoning Primitives
% Numbers pinned in master Claim 5.

\subsection{Puzzle SFT Induces a Reasoning Vocabulary}
\label{sec:sft-induces}

\begin{figure}[t]
  \centering
  \resizebox{0.75\textwidth}{!}{%
    \includegraphics[width=\textwidth]{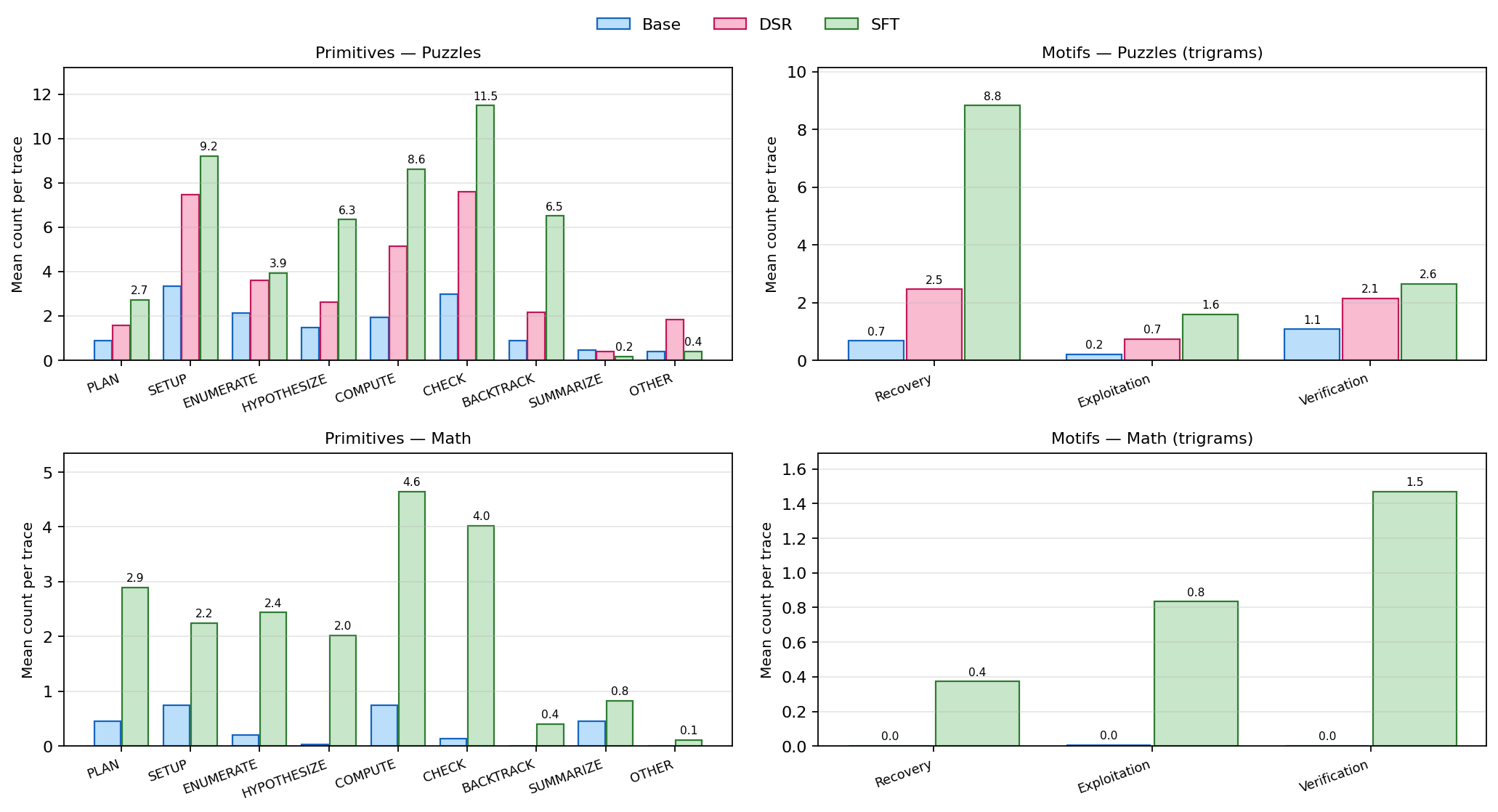}%
  }
  \caption{Per-trace primitive counts (left) and trigram-motif counts (right)
    after puzzle SFT. The top row compares base, DSR-teacher, and post-SFT
    rollouts on held-out puzzles; the bottom row compares base and post-SFT
    rollouts on OlymMATH-Hard. DSR is absent from the math row because it was
    not evaluated on math. Motif definitions are in
    Appendix~\ref{app:motif-details}.}
  \label{fig:primitive-motif-distributions}
\end{figure}

Before any RL, we ask what puzzle SFT alone contributes to reasoning behaviour.
Applying our primitive classifier (Section~\ref{sec:framework}), we compare
base, DSR-teacher, and post-SFT rollouts on held-out puzzles, and base versus
post-SFT rollouts on OlymMATH-Hard. If puzzle-only SFT transfers, the math
rollouts should acquire the same primitive vocabulary even though no mathematics
appears in the SFT corpus. Figure~\ref{fig:primitive-motif-distributions}
tests this at both the primitive and trigram-motif levels.

\paragraph{Puzzle SFT induces the vocabulary used later.}
On held-out puzzles, post-SFT rollouts broadly inherit the DSR teacher's
primitives and motif composition, showing that puzzle SFT induces the teacher's reasoning
style in addition to improving test accuracy. The recovery primitives
\textsc{hypothesize} and \textsc{backtrack} are amplified relative
to the base model, while the main exploitation primitives remain close to the
teacher distribution (Appendix~\ref{app:sft-puzzle-primitives}).

The transfer tests address whether this vocabulary appears out-of-domain. On
OlymMATH-Hard, the base model has little structured primitive vocabulary:
only $2.8$ primitive spans per trace, with essentially no
\textsc{hypothesize} or \textsc{backtrack}. After puzzle SFT, aggregate
primitive count rises to $19.6$, \textsc{check} grows from $0.1$ to $4.0$, and
\textsc{hypothesize}/\textsc{backtrack} appear at $2.0$ and $0.4$ episodes per
trace. Since the SFT corpus contains only puzzle traces, these changes indicate
that puzzle SFT induces a reusable reasoning vocabulary. At the motif level,
recovery, exploitation, and verification trigrams are likewise introduced by
SFT (definitions in Appendix~\ref{app:motif-details}), establishing the
reasoning vocabulary that the following sections examine under RL.

Having established that puzzle SFT makes these primitives available, we next ask
how vanilla RL reallocates them.

% Section 6: What Vanilla RL Changes — Deeper Compute-Verify Chains
% Compact rewrite in progress. Starts from 06_rl_deepens.tex and tightens
% paragraph-by-paragraph.

\subsection{What Vanilla RL Changes: Deeper Compute--Verify Chains}
\label{sec:rl-deepens}

What does vanilla RL change relative to puzzle SFT? Our answer is not that it
discovers a new reasoning vocabulary. The SFT model already has the relevant
primitives; vanilla GSPO instead changes how strongly the model commits to one
part of that vocabulary. On OlymMATH-Hard, the dominant shift is toward deeper
exploitation: longer contiguous runs of \textsc{compute}, \textsc{check}, and
\textsc{setup} episodes that carry an approach forward once the model is on
track. The same depth allocation also appears on harder puzzles, paralleling
the math-side shift.

\begin{figure}[t]
  \centering
  \begin{minipage}{0.30\linewidth}
    \centering
    \includegraphics[width=\linewidth]{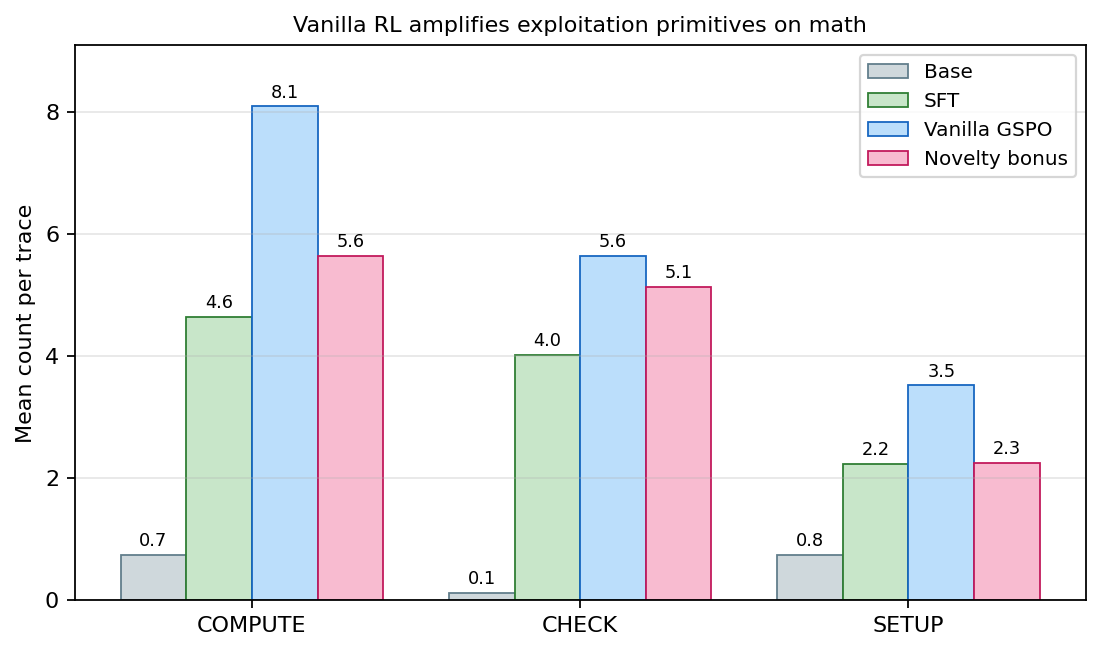}
    \phantomsubcaption
    \label{fig:exploit-primitives}
  \end{minipage}\hfill
  \begin{minipage}{0.68\linewidth}
    \centering
    \includegraphics[width=\linewidth]{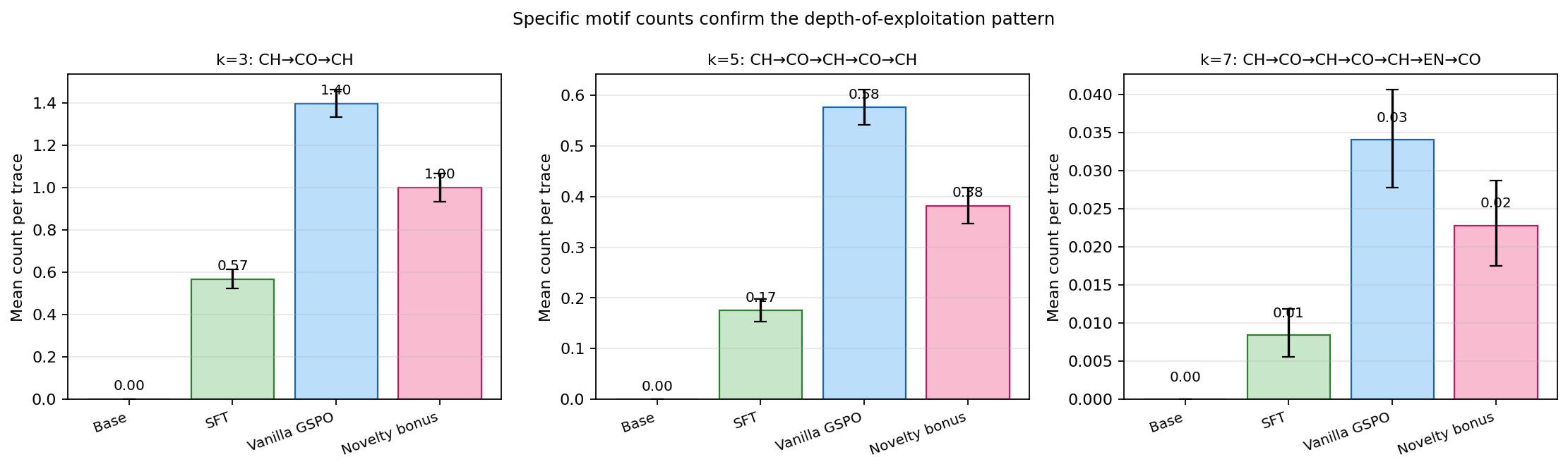}
    \phantomsubcaption
    \label{fig:motif-examples}
  \end{minipage}
  \caption{The SFT$\to$vanilla-GSPO transition amplifies exploitation on
    OlymMATH-Hard. \textbf{(a)} \textsc{compute}, \textsc{check}, and
    \textsc{setup} rise under vanilla GSPO, with a modest reduction under the
    novelty bonus. \textbf{(b)} The same shift appears in $k$-gram motifs,
    especially longer compute--verify chains; error bars are $95\%$ CI.}
  \label{fig:exploit-primitives-and-motifs}
\end{figure}

\paragraph{From primitives to execution chains.}
At the primitive level, the three exploitation primitives all rise on
OlymMATH-Hard (Figure~\ref{fig:exploit-primitives-and-motifs}(a)):
\textsc{compute} nearly doubles ($4.6 \to 8.1$ episodes per trace),
\textsc{check} grows from $4.0$ to $5.6$, and \textsc{setup} from $2.2$ to
$3.5$. The same shift appears in concrete motifs
(Figure~\ref{fig:exploit-primitives-and-motifs}(b)). Short compute--verify
patterns increase, but the larger effect is on longer alternations such as
\textsc{check}$\to$\textsc{compute}$\to$\textsc{check}$\to$\textsc{compute}$\to$\textsc{check},
which rises $3.4\times$ from SFT to vanilla GSPO. This is the first sign that
vanilla RL is not merely producing more local verification; it is assembling
longer execution chains from the same primitives SFT introduced.

\begin{figure}[t]
  \centering
  \begin{minipage}{0.45\linewidth}
    \centering
    \includegraphics[width=\linewidth]{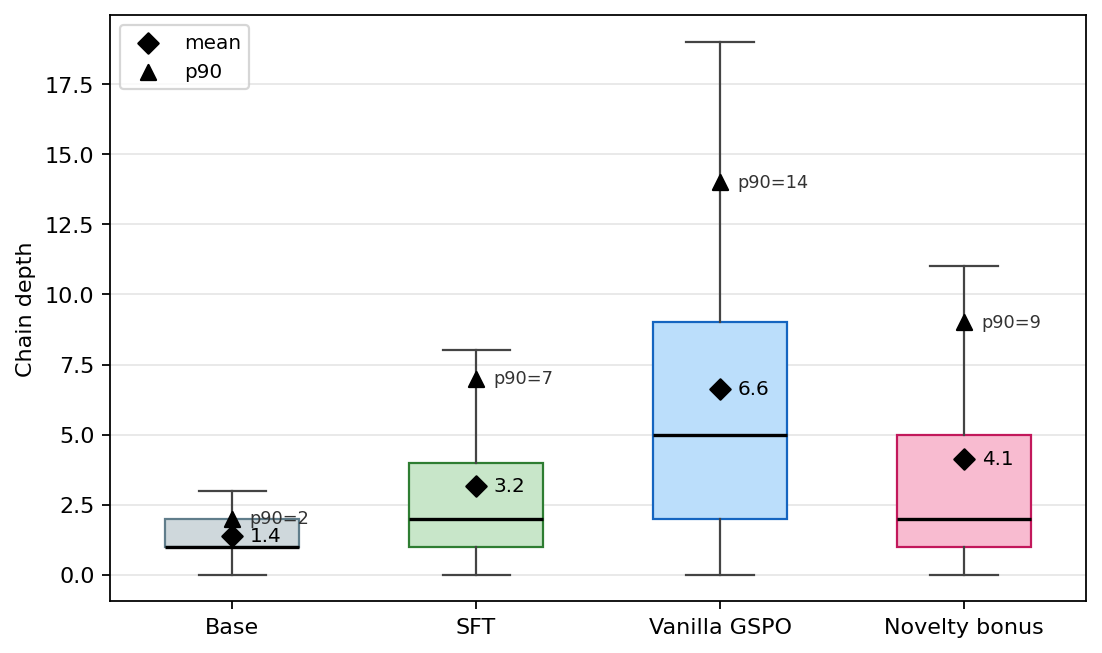}
    \phantomsubcaption
    \label{fig:chain-depth}
  \end{minipage}\hfill
  \begin{minipage}{0.53\linewidth}
    \centering
    \includegraphics[width=\linewidth]{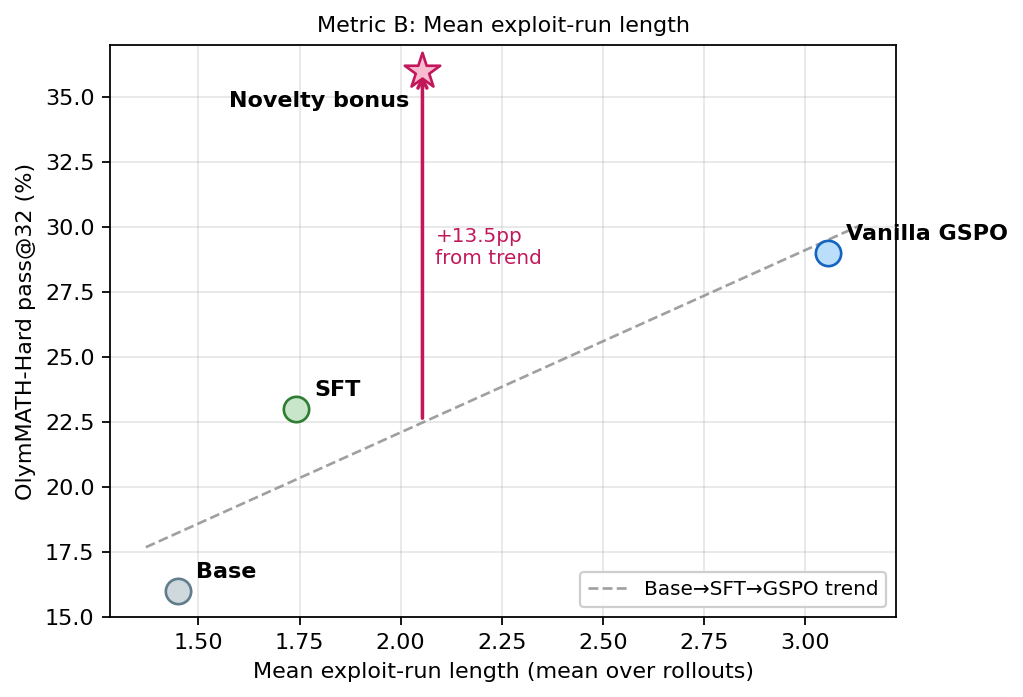}
    \phantomsubcaption
    \label{fig:run-length-vs-passk}
  \end{minipage}
  \caption{Depth statistics on OlymMATH-Hard. \textbf{(a)} Vanilla GSPO
    roughly doubles mean chain depth and raises the upper tail relative to SFT.
    \textbf{(b)} Mean exploit-run length increases along the
    Base$\to$SFT$\to$vanilla-GSPO trajectory, while the novelty-bonus variant
    achieves higher \texttt{pass@32} with shorter exploit runs.}
  \label{fig:depth-summary}
\end{figure}

\paragraph{A parallel depth shift on hard puzzles.}
Frequency alone does not tell us whether these primitives are chained together,
so we next measure depth directly. We define chain depth as the longest
contiguous run of exploitation episodes in a trace before any other primitive
intervenes (Appendix~\ref{app:metrics}). On OlymMATH-Hard, mean chain depth doubles from SFT to vanilla GSPO
($3.2 \to 6.6$), and the $90$-th percentile rises from $7$ to $14$
(Figure~\ref{fig:depth-summary}(a)). The same depth allocation appears on puzzles:
holding the checkpoint fixed, chain depth increases on harder puzzle types and
larger grids, and vanilla GSPO allocates depth more aggressively than SFT
(Appendix~\ref{app:within-checkpoint-depth}). To check that this is not driven
by a few exceptional traces, we also measure the mean length of all exploit runs
per trace; that measure rises from $1.74$ after SFT to $3.06$ after vanilla GSPO
(Figure~\ref{fig:depth-summary}(b)). Thus vanilla RL learns extended compute--verify execution on harder puzzle
instances.

On OlymMATH-Hard, this depth shift tracks the vanilla-RL gain:
\texttt{pass@32} rises from $23.0\%$ at SFT to
$29.0\%$, a $+6$pp gain that accompanies the depth doubling above. This gain
represents a ceiling extension --- in contrast to the convergence
pattern of \citet{yue2025limitrlvr}, puzzle RL adds genuinely new solvable
problems rather than rearranging the SFT distribution. Along the
Base$\to$SFT$\to$vanilla-GSPO trajectory, mean exploit-run length and
\texttt{pass@32} rise together (Figure~\ref{fig:depth-summary}(b)). Paired
inspection of SFT and vanilla-GSPO rollouts on problems solved by both
checkpoints gives the same interpretation: vanilla GSPO usually improves
\emph{execution} of an approach rather than changing the \emph{set} of
approaches attempted (Appendix~\ref{app:cot-inspection}, Part~A). The novelty-bonus variant later breaks this depth trend, gaining
another $+7$pp \texttt{pass@32} with shorter exploit runs; the next section
shows what vanilla RL suppresses.

% Section 7: What Vanilla RL Suppresses
% Compact rewrite in progress. Starts from 07_rl_collapses.tex and tightens
% paragraph-by-paragraph.

\subsection{What Vanilla RL Suppresses}
\label{sec:rl-collapses}

\begin{figure}[t]
  \centering
  \begin{minipage}{0.48\linewidth}
    \centering
    \includegraphics[width=\linewidth]{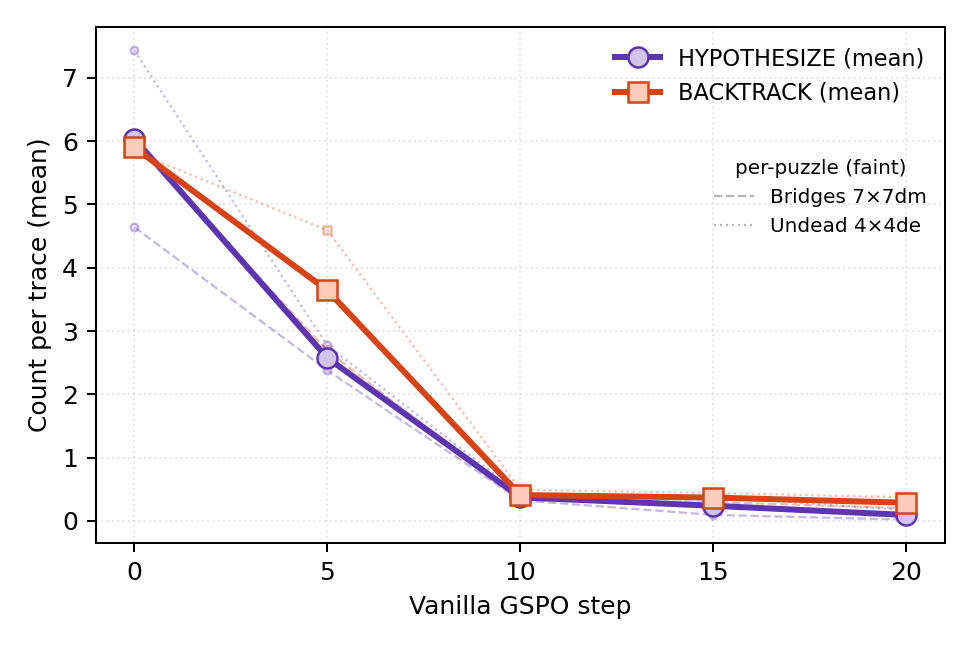}
    \phantomsubcaption
    \label{fig:exploration-counts-puzzles}
  \end{minipage}\hfill
  \begin{minipage}{0.50\linewidth}
    \centering
    \includegraphics[width=\linewidth]{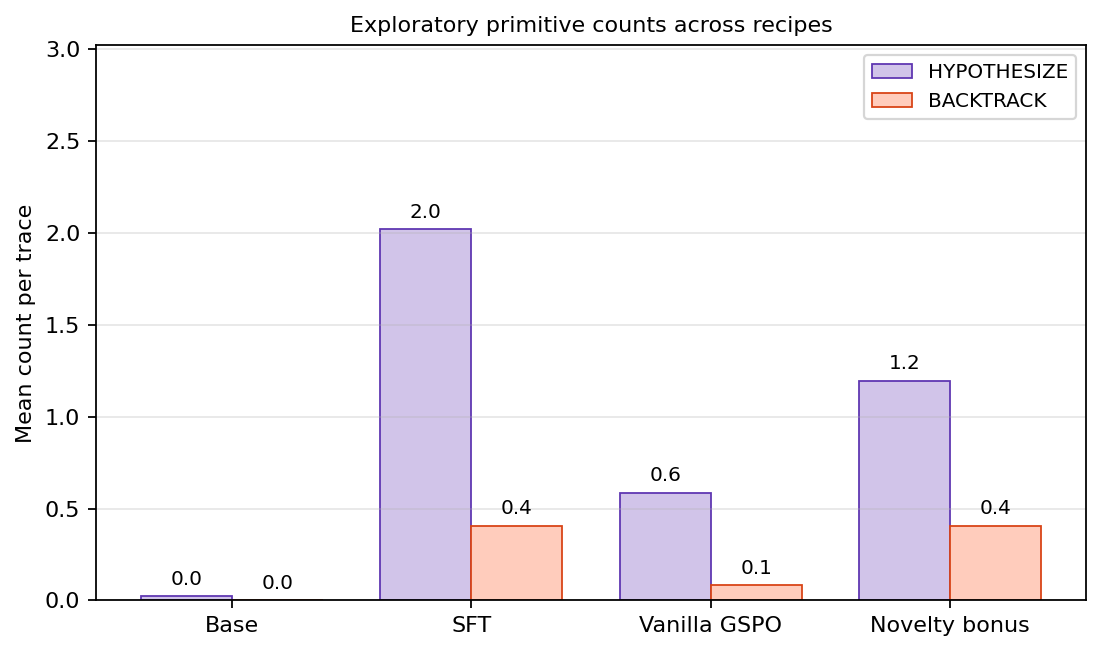}
    \phantomsubcaption
    \label{fig:exploration-counts}
  \end{minipage}
  \caption{Vanilla RL suppresses \textsc{hypothesize} and \textsc{backtrack}.
    \textbf{(a)} On the puzzle training distribution, both primitives fall
    across vanilla-GSPO training as puzzle pass@$k$ rises. \textbf{(b)} The same
    suppression appears on OlymMATH-Hard; the novelty bonus preserves both
    primitives closer to their SFT levels.}
  \label{fig:exploration-suppression}
\end{figure}

The depth gain in Section~\ref{sec:rl-deepens} comes with a cost. As vanilla
GSPO commits more strongly to compute--verify execution, it suppresses the
exploratory primitives that puzzle SFT introduced. On OlymMATH-Hard,
\textsc{hypothesize} falls from $2.0$ to $0.6$ episodes per trace ($-70\%$),
and \textsc{backtrack} from $0.41$ to $0.08$ ($-80\%$). On the puzzle training
distribution itself the suppression is sharper and visibly developmental:
across $20$ vanilla-GSPO steps, both primitives fall monotonically by
$95$--$98\%$ while puzzle solve rates rise substantially
(Figure~\ref{fig:exploration-suppression}(a)). The novelty bonus restores both
primitives toward their SFT levels on math
(Figure~\ref{fig:exploration-suppression}(b)). Vanilla RL therefore does not
merely deepen exploitation; it also prunes the recovery vocabulary that would
let the model recover when a direct line of reasoning fails.

\paragraph{Why does vanilla RL suppress its own exploratory vocabulary?}
A simple explanation is that, on the puzzle training distribution, vanilla RL
makes the model better at recognizing and executing the right approach. Once the
policy can usually identify a productive route, the highest-reward behavior is
not to search over alternatives; it is to carry that route farther through
setup, computation, and checking --- the same pressure that produces the depth
increase of Section~\ref{sec:rl-deepens}. In that regime, \textsc{hypothesize} and
\textsc{backtrack} introduce detours without often changing the outcome, so
continued on-policy training can make them less likely. Figure~\ref{fig:exploration-suppression}(a) is consistent with this
interpretation: as puzzle accuracy rises, the policy gives up the recovery
moves that would help when its first approach is wrong.

\paragraph{But hard math is not the training distribution.}
That tradeoff is useful only when the first approach is usually right. Hard math
often violates this assumption: a promising route may lead to a dead end, and
solving requires trying another representation, case split, or construction. A
model whose \textsc{hypothesize} and \textsc{backtrack} primitives have been
suppressed can still execute a chosen approach deeply, but it is less able to
recover when that approach fails. The novelty GSPO variant, also trained from the SFT checkpoint, preserves
both primitives toward their SFT levels on math and raises OlymMATH-Hard
\texttt{pass@32} to $36.0\%$, compared with $29.0\%$ for vanilla GSPO and
$23.0\%$ for SFT
(Table~\ref{tab:main}); Section~\ref{sec:confirmation} shows that the gain is
concentrated on problems whose solved traces use exactly this recovered
structure.

% Note: the "alternative we tried (entropy bonus)" paragraph that lived here was
% moved to Section~\ref{sec:novelty} (intervention) where the alternative-method
% comparison naturally belongs alongside the controlled A/B.

Section~\ref{sec:novelty} introduces a targeted intervention designed to preserve
the recovery primitives this section showed vanilla RL suppresses.

% Section 8: Intervention — Frozen-Reference Novelty Bonus
% Compact rewrite in progress. Starts from 08_novelty.tex and tightens
% paragraph-by-paragraph.

\section{A Frozen-Reference Novelty Bonus Preserves Recovery}
\label{sec:novelty}

Sections~\ref{sec:rl-deepens} and \ref{sec:rl-collapses} give us a design target:
preserve the recovery primitives that vanilla RL suppresses without giving up
the compute--verify depth it learns. We do this by adding a reward bonus for
correct rollouts whose reasoning is unusually surprising under the frozen SFT
reference model. The bonus is applied only to verifier-correct rollouts and normalised within
the correct rollouts on the same prompt: it rewards alternative successful
reasoning paths, not arbitrary low-probability text.

\paragraph{Bonus definition.}
For each rollout $r$ on prompt $p$, we evaluate the generated tokens under the
frozen SFT reference $\pi_{\mathrm{SFT}}$, with token loss
$\mathrm{NLL}_t(r)=-\log \pi_{\mathrm{SFT}}(r_t \mid r_{<t},p)$. We define the
rollout novelty score as the mean of the top-$k$ largest token NLLs:
\begin{equation}
  s(r) = \frac{1}{k} \sum_{t \in \mathrm{top}_k(\mathrm{NLL}(r))} \mathrm{NLL}_t(r),
  \qquad k = 100.
\end{equation}
For correct rollouts within each prompt group, we $z$-score and clip the
novelty scores before adding them to the verifier reward:
\begin{equation}
  z_p(r) = \mathrm{clip}\!\left(\frac{s(r) - \mu_p}{\sigma_p},\; -2,\; 2\right),
  \qquad
  R_{\mathrm{novelty}}(r)
  =
  R_{\mathrm{base}}(r)
  +
  \alpha z_p(r)\mathbb{1}[r \text{ correct}].
\end{equation}
where $\mu_p$ and $\sigma_p$ are computed over correct rollouts for prompt $p$.
We set $\alpha=0.1$. Thus incorrect rollouts receive no novelty credit, and
correct rollouts receive more credit when their reasoning is more novel
relative to the SFT reference than other correct rollouts on the same prompt. Pseudocode is in Appendix~\ref{app:novelty-pseudocode}.

\paragraph{Why frozen-reference top-$k$ NLL?}
The frozen reference keeps the novelty target stable. If novelty were ranked
under the current policy, as in unlikeliness-style rewards
\citep{he2025unlikeliness}, the target would drift toward whatever the model is
already doing well, weakening the diversity signal as training progresses. The
top-$k$ statistic makes the signal local enough to matter at our trace lengths:
with $\sim$25k-token rollouts, mean NLL has near-zero within-group variance
because most tokens are high-probability formatting or structural choices. Using
only the $k=100$ highest-NLL tokens isolates rare decision points and amplifies
the within-group signal by $10$--$15\times$
(Appendix~\ref{app:novelty-signal-analysis}).

\paragraph{Effect.}
The novelty-bonus variant starts from the same SFT model as vanilla GSPO and
uses the same data, algorithm, training budget, and number of optimisation steps. It
raises OlymMATH-Hard \texttt{pass@32} to $\mathbf{36.0\%}$, compared with
$29.0\%$ for vanilla GSPO and $23.0\%$ for the shared SFT starting point
(Table~\ref{tab:main}; Figure~\ref{fig:main-passk-hard}). Mechanistically, it
restores the recovery primitives that vanilla GSPO suppressed:
\textsc{backtrack} returns to its SFT level and \textsc{hypothesize} recovers
to about $60\%$ of its SFT rate (Figure~\ref{fig:exploration-counts}). This
restoration comes with less exploitation depth: chain depth, exploit-run
length, and the exploitation motifs amplified by vanilla GSPO all reduce modestly
(Figures~\ref{fig:depth-summary}(b) and
\ref{fig:exploit-primitives-and-motifs}(b)). The bonus therefore does not
improve by adding still more depth; it trades some direct compute--verify
execution for recovered ability to try and abandon alternatives.

We also screened diversity-promoting alternatives in smaller pilots, including
a loss-level entropy bonus and entropy-advantage variants inspired by
\citet{cheng2025reasoningexploration}. These pilot experiments were used for method
selection, not as a definitive full-scale comparison: entropy matched the
novelty bonus at mini scale but was less stable, while entropy-advantage
variants collapsed below baseline. We therefore scaled the frozen-reference
novelty bonus for the GSPO comparison; the pilot summary and its limitations are
in Appendix~\ref{app:diversity-alternatives}. The next section examines whether the restored recovery primitives account
for the performance gain.

% Section 9: Mechanism Confirmation — recovery on unique-to-novelty problems
% Compact rewrite in progress. Starts from 09_confirmation.tex and tightens
% paragraph-by-paragraph.

\subsection{Where the Novelty Gain Comes From: Recovery on Unique-to-Novelty Problems}
\label{sec:confirmation}

Section~\ref{sec:novelty} left a specific question: does the novelty bonus
restore \textsc{hypothesize}/\textsc{backtrack} everywhere, or does it deploy
recovery selectively on the problems where direct compute--verify execution
fails? We test this on the $100$ OlymMATH-Hard problems by comparing two diagnostic
subsets: problems solved by both vanilla GSPO and the novelty-bonus variant
(``shared''), and problems solved only by the novelty-bonus variant
(``unique-to-novelty''). The full accounting also includes vanilla-only and
unsolved problems (Appendix~\ref{app:per-problem-recovery-split}), but the
contrast between shared and unique-to-novelty problems is the cleanest test of
selectivity: shared problems show whether the novelty-bonus variant behaves
like vanilla GSPO when direct exploitation is enough, while unique-to-novelty
problems show whether recovery appears where vanilla GSPO fails.

\begin{figure}[t]
  \centering
  \begin{subfigure}{0.66\linewidth}
    \centering
    \includegraphics[width=\linewidth]{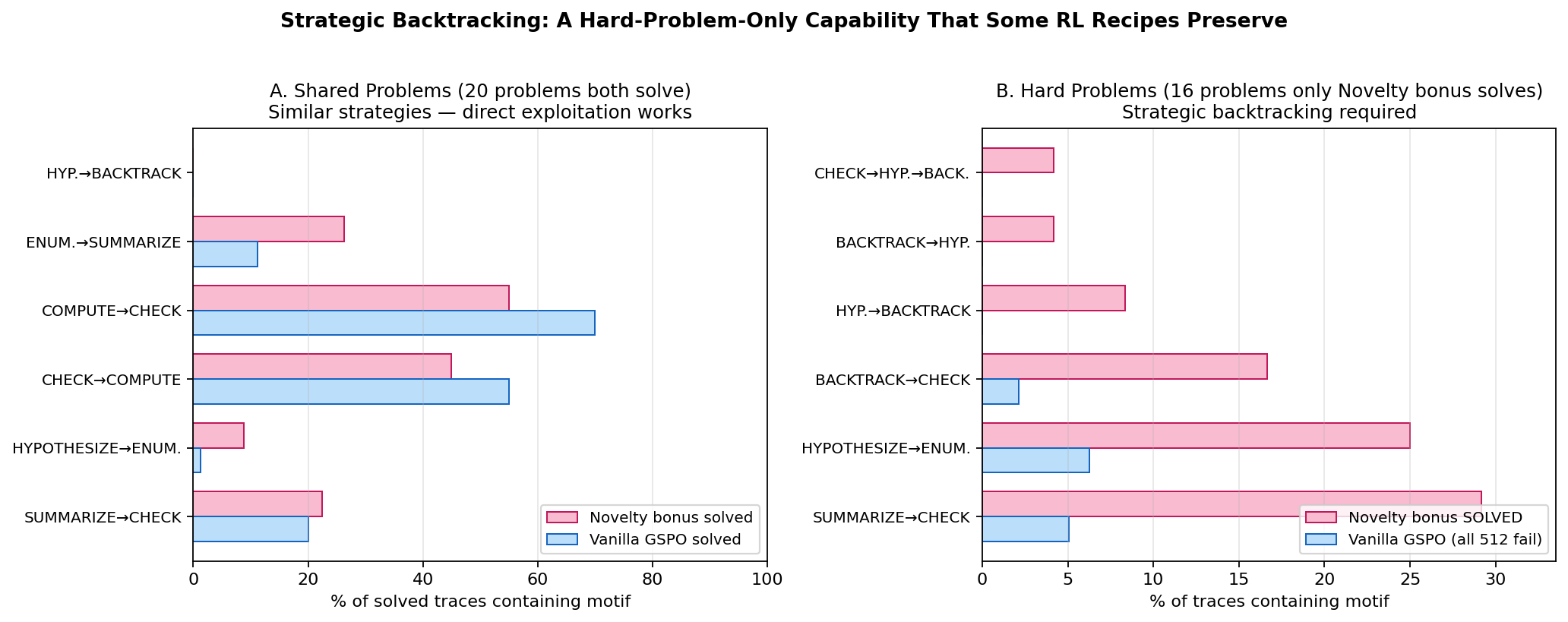}
    \caption{Recovery motifs by problem subset}
    \label{fig:confirmation-motifs}
  \end{subfigure}\hfill
  \begin{subfigure}{0.31\linewidth}
    \centering
    \includegraphics[width=\linewidth]{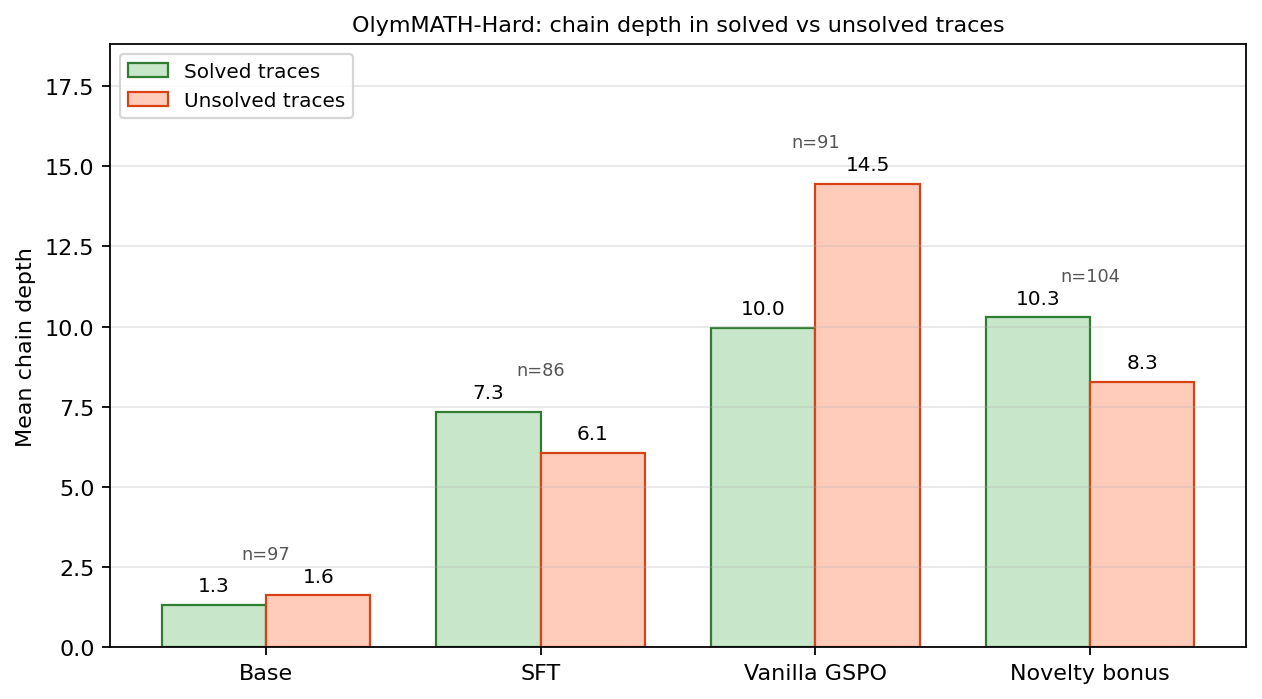}
    \caption{Depth split by trace success}
    \label{fig:confirmation-depth}
  \end{subfigure}
  \caption{Diagnostic evidence for the novelty gain on OlymMATH-Hard.
    \textbf{(a)} On shared problems, vanilla GSPO and novelty rollouts have
    similar motif distributions; on unique-to-novelty problems, recovery motifs
    remain available only under the novelty bonus. \textbf{(b)} Vanilla GSPO
    spends more depth on failures, while the novelty-bonus variant reverses
    the sign; significance tests are in Table~\ref{tab:chain-depth-sig}.}
  \label{fig:confirmation-evidence}
\end{figure}

\paragraph{Recovery motifs are selective, not uniform.}
The aggregate motif analysis shows where the novelty bonus changes behavior
(Figure~\ref{fig:confirmation-evidence}(a)). On the $20$ shared problems, solved traces from
vanilla GSPO and the novelty-bonus variant look nearly identical: both rely on
direct \textsc{compute}$\to$\textsc{check} exploitation, and recovery motifs
remain rare. On the $16$ unique-to-novelty
problems, the pattern changes. The recovery motifs are absent from
vanilla GSPO's $512$ attempts but remain available in the novelty-bonus variant
($11/512$ rollouts). Within the novelty-bonus run, they are also enriched in
successful traces: $3/24$ solved traces contain at least one diagnostic recovery
motif, compared with $8/488$ failed traces, an approximately $8\times$
difference. Thus the gain is not explained by recovery being used uniformly
across novelty rollouts. Instead, the bonus preserves a recovery move that
vanilla GSPO has lost, and that move appears disproportionately in successful
unique-to-novelty traces. This is the selectivity pattern Section~\ref{sec:rl-collapses} predicted:
recovery is unnecessary on shared problems but most useful on problems where
vanilla GSPO's direct execution fails. Appendix~\ref{app:cot-inspection}, Part~B, gives two full trace
pairs showing the same pivot in text.

\paragraph{Novelty aligns depth with successful trajectories.}
Section~\ref{sec:rl-deepens} left a puzzle: the novelty-bonus variant
achieves higher \texttt{pass@32} than vanilla GSPO despite shorter average
exploit runs. Splitting traces by whether they solve the problem clarifies this
pattern. Vanilla GSPO is the only RL run whose unsolved traces are deeper than
its solved ones; the novelty-bonus variant reverses the sign, with solved traces
deeper than unsolved (Figure~\ref{fig:confirmation-evidence}(b)). This
suggests that vanilla GSPO often spends depth on unsuccessful continuations,
whereas the novelty bonus makes depth more coupled to trajectories that succeed.

% Section 10: Conclusion and Limitations

\section{Conclusion and Limitations}
\label{sec:discussion}

We post-trained a 7B model with no mathematics in any post-training stage and
found that puzzle-only SFT and RL can raise the sampled ceiling on hard
mathematics. The primitive
analysis suggests a more specific lesson than ``more reasoning helps.'' Puzzle
SFT makes a reusable vocabulary of planning, checking, hypothesizing, and
backtracking available out of domain; vanilla GSPO then sharpens execution by
lengthening compute--verify chains, but at the cost of suppressing recovery
moves. The frozen-reference novelty bonus improves transfer by preserving some
of that recovery vocabulary while retaining enough exploitation depth. More
broadly, cross-domain RL transfer appears to depend not only on how far a model
can pursue an approach, but also on whether training preserves the ability to
abandon that approach when it fails. Taken together, our contribution is an empirical test of whether RL-trained
LLMs can turn a shared problem-solving vocabulary into measurable
out-of-domain math gains.

\paragraph{Limitations.}
Our results are limited to one base model (OLMo3-7B-Instruct-SFT) and one RL
family (GSPO); the novelty-bonus hyperparameters were selected from pilot runs
rather than exhaustive tuning; and the primitive analysis is a diagnostic lens,
not a per-problem causal proof. Extended discussion is in
Appendix~\ref{app:limitations}.

\bibliographystyle{plainnat}
\bibliography{references,cogsci}

%%%%%%%%%%%%%%%%%%%%%%%%%%%%%%%%%%%%%%%%%%%%%%%%%%%%%%%%%%%%

\appendix

\section{Broader impact}
\label{app:broader-impact}

Our contribution improves out-of-domain reasoning of a small open research
model via verifier-only puzzle RL; the standard dual-use considerations for
reasoning LLMs apply, and the method does not introduce capability surfaces
beyond those already accessible from the base model.

\section{Compute resources}
\label{app:compute}

Stage~1 SFT runs on $8\times$B200 (80\,GB) for $\approx$9\,hr. The two
Stage~2 RL variants---vanilla GSPO and the novelty-bonus GSPO---are
parallel branches from the same SFT checkpoint and each runs on $8\times$B200
for $\approx$70\,hr. End-to-end reproduction of the headline OlymMATH-Hard
$36.0\%$ pass@32 number (SFT plus the novelty-bonus branch) is therefore
approximately $632$ GPU-hours; reproducing both Stage~2 branches together
is approximately $1{,}192$ GPU-hours. The full research project---preliminary
pilot runs, abandoned hyperparameter configurations, and ablations not
reported in the paper---required roughly an additional $2$--$3\times$ the
reproduction budget on top.

\section{Limitations and future work}
\label{app:limitations}

Our evidence comes from one base model, OLMo3-7B-Instruct-SFT, and one RL
algorithm family, GSPO. Because full-scale runs with long puzzle rollouts are
computationally expensive, we leave validation on other models, scales, and RL
objectives to future work. The novelty bonus also has design choices that were
not exhaustively tuned at full scale: $\alpha = 0.1$ and top-$k = 100$ were
selected from smaller pilot runs. For the same reason, alternative diversity
mechanisms were screened in smaller pilots rather than compared at the full
training scale. Finally, our primitive analysis is a diagnostic lens, not a
causal proof for every solved problem: the classifier was validated on this model
family and trace style, and the motif evidence shows where recovery is available
and enriched, not that every novelty-only solution uses it.

\section{Detailed accuracy tables}
\label{app:result-tables}

This appendix collects the per-benchmark accuracy tables referenced from
Section~\ref{sec:results}. Table~\ref{tab:main-puzzles} reports in-domain
puzzle accuracy at \texttt{pass@1}; Table~\ref{tab:main} reports held-out
mathematics at \texttt{pass@1} and \texttt{pass@32}. Both are restated here
for ease of cross-reference; Figure~\ref{fig:math-passk} summarises the same
data as \texttt{pass@}$k$ curves for the body benchmarks.

% --- Puzzle in-domain results table (pass@1 only; pass@32 saturates for SFT) ---
\begin{table}[h]
  \caption{In-domain puzzle accuracy on held-out test sets at training grid sizes,
    \texttt{pass@1} (greedy, $200$ problems per grid). We omit \texttt{pass@32}: it
    saturates for the SFT baseline and does not differentiate the recipe stages.
    Each RL row reports the run's best-puzzle-aggregate step. The novelty bonus's
    puzzle peak (step 10) precedes its math peak (step 15, the headline math checkpoint
    in Table~\ref{tab:main}); we report both for transparency.}
  \label{tab:main-puzzles}
  \centering
  \small
  \resizebox{\linewidth}{!}{%
  \begin{tabular}{lcccc}
    \toprule
    & Bridges $7{\times}7$ & Pattern $4{\times}4$ & Undead $4{\times}4$ & Galaxies $3{\times}3$ \\
    \midrule
    Base                                                  & 0.0   & 0.0   & 0.0   & 0.0   \\
    + Puzzle SFT (rejection-sampled DSR)                  & 28.0  & 34.3  & 22.5  & 45.0  \\
    \textbf{+ GSPO RL (vanilla; ours)}                    & \textbf{79.5} & 53.5 & \textbf{57.0} & 65.0 \\
    + GSPO RL with novelty bonus (step 10, best puzzle)   & 50.0 & 56.6 & 47.0 & \textbf{80.0} \\
    + GSPO RL with novelty bonus (step 15, best math; ours) & 46.0 & \textbf{66.7} & 47.0 & 67.5 \\
    \bottomrule
  \end{tabular}}
\end{table}

% --- Math results table (pass@1 + pass@32) ---
\begin{table}[h]
  \caption{Capability on held-out mathematics. The recipe trains on constraint-satisfaction
    puzzles only; OlymMATH and AIME25 never appear in training. We report \texttt{pass@1}
    (accuracy) and \texttt{pass@32} (capability ceiling) over 32 rollouts per problem at
    temperature 0.6. OlymMATH-Hard is the difficulty tier of interest: AIME25 and
    OlymMATH-Easy saturate quickly with sampling at this scale, so RL gains there are
    necessarily small. The novelty bonus's largest gain is on the unsaturated benchmark
    (Hard, $+7$pp at \texttt{pass@32}); on every benchmark its effect on \texttt{pass@1} is
    near zero, the signature of a method that preserves multiple correct reasoning paths
    rather than sharpening any single one.}
  \label{tab:main}
  \centering
  \small
  \begin{tabular}{lcccccc}
    \toprule
    & \multicolumn{2}{c}{\textbf{OlymMATH-Hard}} & \multicolumn{2}{c}{AIME25} & \multicolumn{2}{c}{OlymMATH-Easy} \\
    \cmidrule(lr){2-3}\cmidrule(lr){4-5}\cmidrule(lr){6-7}
    & @1 & @32 & @1 & @32 & @1 & @32 \\
    \midrule
    Base (OLMo3-7B-Instruct-SFT)                 & 1.5 & 16.0 & 9.4  & 34.0 & 3.2  & 33.9 \\
    + Puzzle SFT (rejection-sampled DSR traces)  & 2.7 & 23.0 & 27.0 & 73.1 & 14.2 & 72.0 \\
    \textbf{+ GSPO RL (vanilla; ours)}           & 2.8 & 29.0 & \textbf{29.6} & 75.4 & \textbf{15.9} & \textbf{73.0} \\
    \textbf{+ GSPO RL with novelty bonus (ours)} & \textbf{3.2} & \textbf{36.0} & 28.5 & \textbf{76.7} & 15.4 & 71.0 \\
    \bottomrule
  \end{tabular}
\end{table}

\begin{figure}[t]
  \centering
  \includegraphics[width=\linewidth]{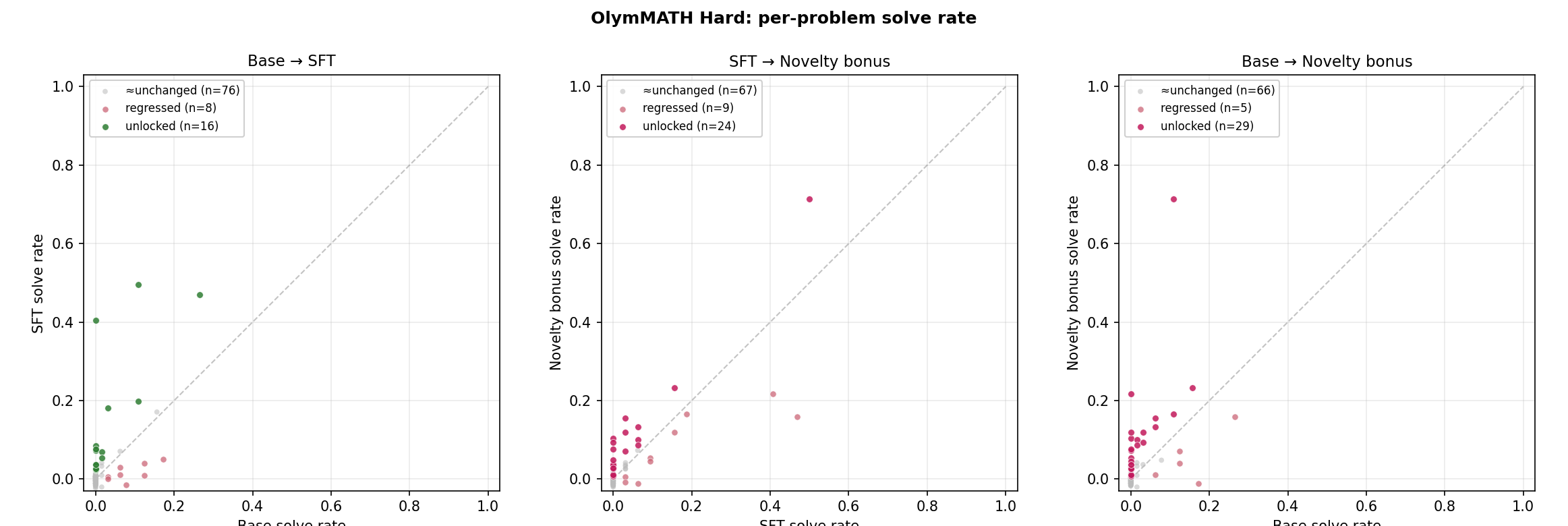}
  \caption{Per-problem solve rates on OlymMATH-Hard ($100$ problems, $32$
    rollouts per problem) across the recipe. Each point is one problem;
    colour marks problems whose solve rate changed by more than the
    pass-rate noise floor. The primary \texttt{pass@32} ceiling rise is
    driven by genuine per-problem unlocks: Base$\to$Novelty newly solves
    $29$ problems and regresses $5$ ($\sim$$6{:}1$), with most of the unlocks
    happening past SFT (SFT$\to$Novelty unlocks $24$, regresses $9$). The
    recipe lifts the ceiling on a broad set of problems rather than
    sharpening sampling on a few already-solved ones.}
  \label{fig:per-problem-progression}
\end{figure}

\section{Additional benchmark \texttt{pass@}$k$ curves}
\label{app:additional-benchmarks}

Figure~\ref{fig:math-passk} (body) reports OlymMATH-Hard, HMMT combined,
OMEGA Explorative, and the held-out larger-grid puzzle evaluations. We document four
additional benchmarks here for completeness: OlymMATH-Easy, AIME25, OMEGA
Compositional, and OMEGA Transformative. None earn a body slot --- each is
either saturated by SFT or has a confound that limits the RL-discriminating
zone --- but together they verify that the recipe does not regress on benchmarks
the body figure does not show.

\begin{figure}[t]
  \centering
  \begin{subfigure}{0.48\linewidth}
    \centering
    \includegraphics[width=\linewidth]{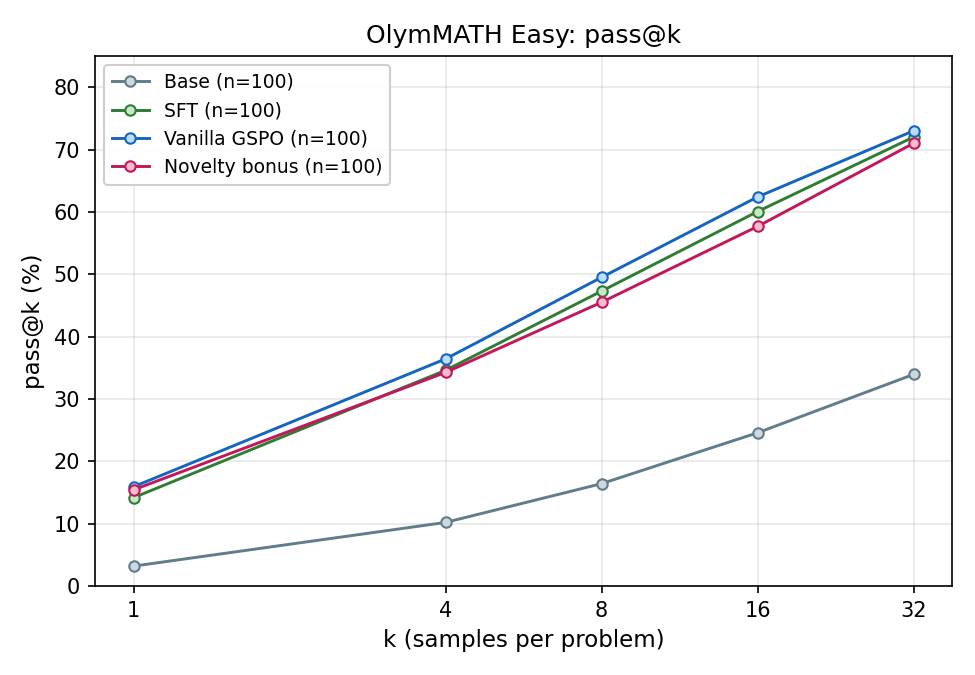}
    \caption{OlymMATH-Easy ($N{=}100$)}
    \label{fig:appendix-passk-easy}
  \end{subfigure}\hfill
  \begin{subfigure}{0.48\linewidth}
    \centering
    \includegraphics[width=\linewidth]{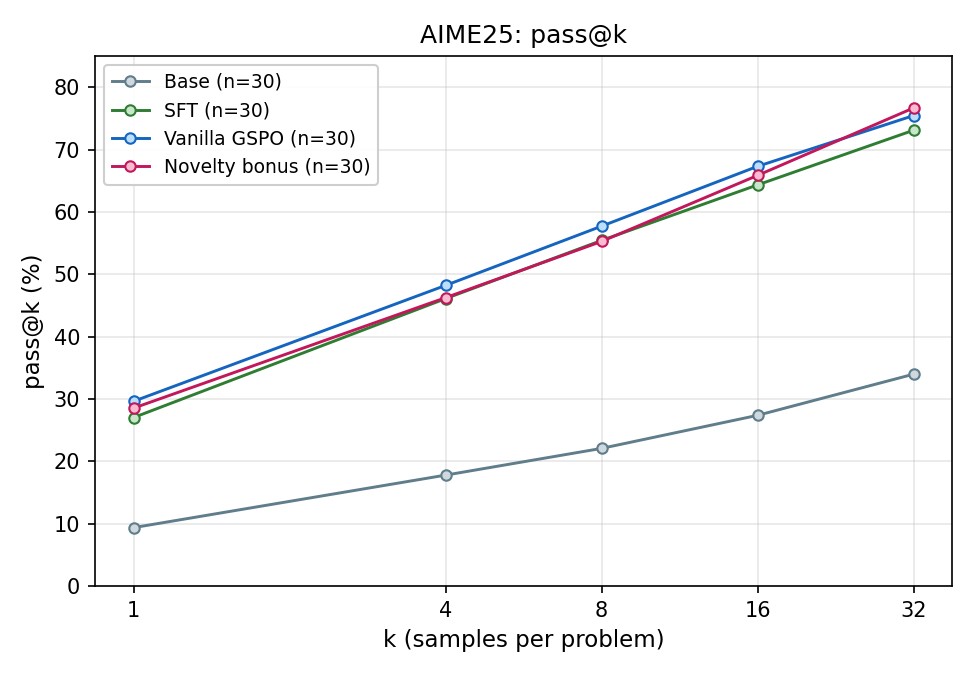}
    \caption{AIME25 ($N{=}30$)}
    \label{fig:appendix-passk-aime25}
  \end{subfigure}\\[6pt]
  \begin{subfigure}{0.48\linewidth}
    \centering
    \includegraphics[width=\linewidth]{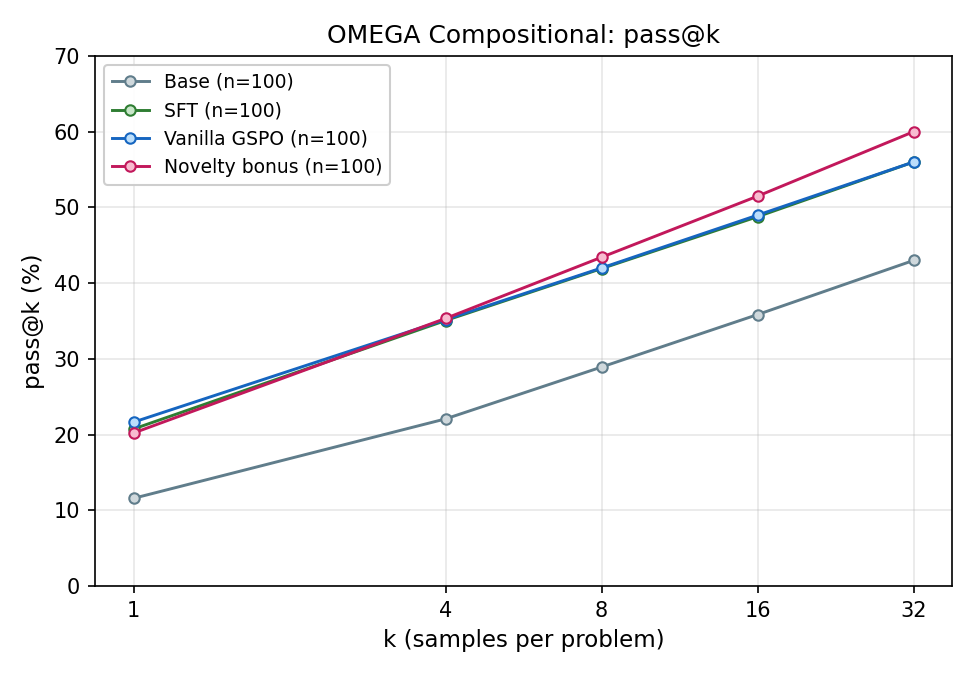}
    \caption{OMEGA Compositional ($N{=}100$)}
    \label{fig:appendix-passk-omega-comp}
  \end{subfigure}\hfill
  \begin{subfigure}{0.48\linewidth}
    \centering
    \includegraphics[width=\linewidth]{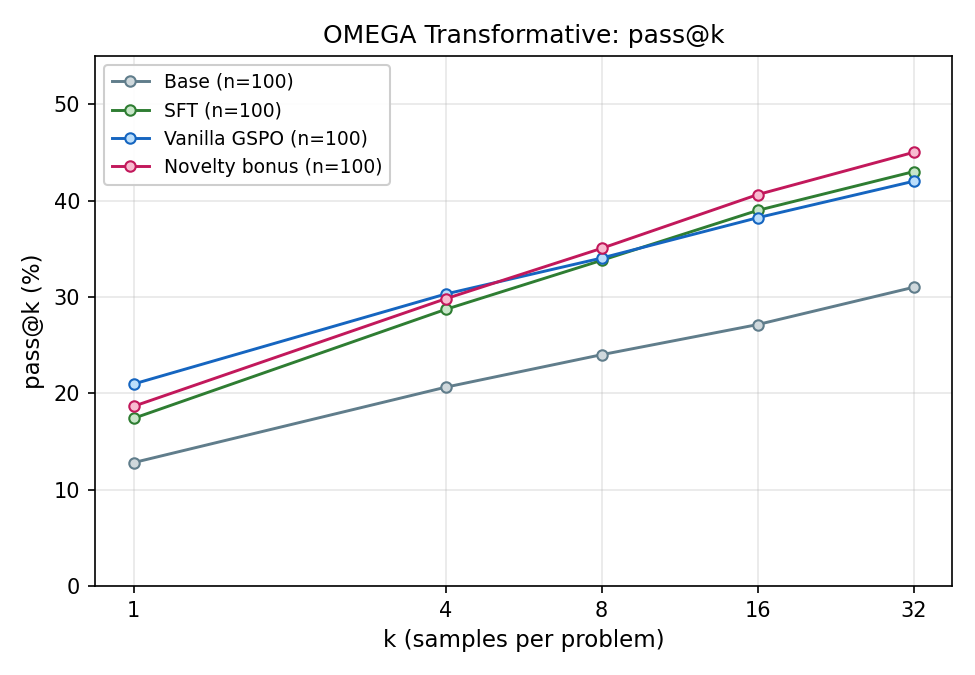}
    \caption{OMEGA Transformative ($N{=}100$)}
    \label{fig:appendix-passk-omega-trans}
  \end{subfigure}
  \caption{Additional math benchmark \texttt{pass@}$k$ curves not shown in the
    body. \textbf{(a)}~OlymMATH-Easy: large Base$\to$SFT lift ($+38$pp pass@$32$,
    $34\to72$); RL contribution is small thereafter as the benchmark
    saturates. \textbf{(b)}~AIME25: same shape as Easy --- Base$\to$SFT carries
    the bulk of the lift ($34\to73$pp), RL stages tied within $\sim 1$--$3$pp
    at every $k$. \textbf{(c)}~OMEGA Compositional ($7$ configs combining
    pairs of math domains): Base$\to$SFT $+12$pp, vanilla GSPO ties SFT,
    novelty adds $+4$pp. SFT already covers the relevant compositional
    skills. \textbf{(d)}~OMEGA Transformative ($7$ configs requiring creative
    reframing): bimodal at $7$B --- $34$ of $100$ problems are universally
    unsolvable (de Moivre, function-intersection), $42$ are Base-saturated,
    only $24$ in the RL-discriminating zone. Recipe is non-monotonic ($31\to
    43 \to 42 \to 45$); vanilla GSPO regresses by $1$pp at pass@$32$, novelty
    recovers and adds $+2$pp over SFT.}
  \label{fig:appendix-passk}
\end{figure}

OlymMATH-Easy and AIME25 saturate quickly with $32$ rollouts at this scale, so
the gap between recipe stages collapses past pass@$8$ --- both Vanilla GSPO and
Novelty are within $\sim 1$--$3$pp of each other and within $\sim 4$pp of SFT
on pass@$32$. They confirm the recipe roughly preserves easier-tier accuracy (differences
between RL stages are within the noise floor at saturation) but
do not differentiate RL stages. OMEGA Compositional shows a near-tie between
SFT and vanilla GSPO at pass@$32$ (both $56\%$); novelty lifts to $60\%$.
OMEGA Transformative is the most stress-test-like split but is bimodal at $7$B
scale: a per-config audit shows only ${\sim}24$ of $100$ problems are in the
RL-discriminating zone --- the rest are either solved by Base ($42$ problems)
or universally unsolvable at this scale ($34$ problems, dominated by de Moivre
and function-intersection configs where every checkpoint scores zero). Vanilla GSPO regresses by a hair on pass@$32$ ($43\to
42$); novelty recovers and adds $+2$pp over SFT.

\paragraph{Evaluation cost note.}
A single pass@$32$ sweep at \texttt{max\_gen\_toks}${=}25{,}000$ generates
$N \times 32$ rollouts of up to $25{,}000$ tokens each. With seed-diversification required for valid pass@$k$ statistics,
the $32$ samples per problem must be sampled independently rather than batched as a fixed-prompt expansion;
the cost is therefore dominated by autoregressive decoding latency, not
forward-pass throughput. On $8{\times}$ B200 with vLLM tensor-parallel${=}1$,
a single $4$-checkpoint sweep at $N{=}100$ takes roughly $11{-}12$ wall-clock
hours; a full $4$-checkpoint sweep at $N{=}300$ would take $\sim 36$ hr. This
is the reason we sub-sample OMEGA splits to $N{=}100$ with a fixed seed for
apples-to-apples scoring across checkpoints; the
sub-sampled estimate is within $\sim 1$pp of the full-set estimate where
verified (OMEGA Explorative sub-100 $72.0\%$ pass@$32$ vs full-set $\sim 72.5\%$).

\section{Per-setting primitive distributions}
\label{app:sft-puzzle-primitives}

We apply our classifier of Section~\ref{sec:framework} to three populations of
reasoning traces and report the relative composition of primitive episodes (percent
of all classified primitive labels). The three settings are: \textbf{DSR puzzles}
(reasoning traces from the DSR teacher on the puzzle problems used in our SFT
corpus); \textbf{SFT puzzles} (post-SFT OLMo3 rollouts on held-out puzzle test sets
at training grid sizes); \textbf{SFT math} (post-SFT OLMo3 rollouts on
OlymMATH-Hard, the same distribution Section~\ref{sec:sft-induces} reports
per-trace counts on).

\begin{table}[h]
  \caption{Relative primitive composition (\% of primitive episodes) across the
    three settings. For \textsc{plan}, \textsc{compute}, \textsc{check}, and
    \textsc{enumerate}, the post-SFT distribution on puzzles closely tracks DSR's
    puzzle distribution, indicating SFT broadly mimics the teacher's reasoning style.
    Recovery primitives (\textsc{hypothesize}, \textsc{backtrack}) are present in
    DSR at $4.6\%$ and $3.2\%$ and amplified post-SFT to $11.0\%$ and $11.3\%$. On
    math (a held-out domain), the SFT distribution shows different relative
    proportions---more \textsc{summarize}, less \textsc{backtrack}---suggesting the
    model deploys reasoning structure conditional on the input domain.}
  \label{tab:per-setting-primitives}
  \centering
  \small
  \begin{tabular}{lccc}
    \toprule
    Primitive & DSR puzzles & SFT puzzles & SFT math \\
    \midrule
    \textsc{plan}        & 4.7  & 5.5  & 16.7 \\
    \textsc{setup}       & 26.8 & 20.3 & 18.9 \\
    \textsc{enumerate}   & 9.6  & 8.4  & 9.1  \\
    \textsc{hypothesize} & 4.6  & 11.0 & 5.4  \\
    \textsc{compute}     & 17.9 & 18.3 & 24.2 \\
    \textsc{check}       & 23.4 & 23.3 & 12.7 \\
    \textsc{backtrack}   & 3.2  & 11.3 & 0.9  \\
    \textsc{summarize}   & 1.6  & 0.5  & 11.8 \\
    \textsc{other}       & 8.2  & 1.5  & 0.3  \\
    \bottomrule
  \end{tabular}
\end{table}

\section{Mechanism patterns across math benchmarks}
\label{app:mechanism-breadth}

The mechanism analysis in Sections~\ref{sec:rl-deepens}--\ref{sec:rl-collapses}
is reported on OlymMATH-Hard. To check that the patterns are not specific to
that benchmark, we run the same classifier on rollouts at the four main
checkpoints on \textbf{OlymMATH-Easy} and \textbf{AIME25} and recompute the
body's load-bearing metrics with the canonical body definitions
(non-overlapping motif counts; chain depth as the longest consecutive run of
\textsc{compute}/\textsc{check}/\textsc{setup}). This appendix reports those
cross-benchmark numbers; the Hard column reproduces the body figures.

The qualitative finding is that every directional ranking in
\S\ref{sec:rl-deepens}--\S\ref{sec:rl-collapses} holds on Easy and AIME25
with similar magnitudes:

\begin{itemize}[leftmargin=*,itemsep=2pt,topsep=4pt]
  \item Chain depth: $\mathrm{Base} < \mathrm{SFT} < \mathrm{Vanilla\ GSPO}$, novelty
    sits between SFT and vanilla GSPO.
  \item \textsc{hypothesize} count: suppressed under vanilla GSPO ($3$--$5\times$
    drop from SFT), partially restored under novelty.
  \item \textsc{backtrack} count: suppressed under vanilla GSPO ($\sim 5\times$),
    fully restored or exceeded by novelty.
  \item $k = 3$ and $k = 5$ exploitation motifs ($\textsc{check}\to\textsc{compute}\to\ldots$):
    rise sharply from SFT to vanilla GSPO, novelty drops back partway.
\end{itemize}

\begin{table}[h]
  \caption{Mean chain depth (longest exploit run per trace).}
  \label{tab:breadth-chain-depth}
  \centering
  \small
  \begin{tabular}{lccc}
    \toprule
    Checkpoint & OlymMATH-Hard & OlymMATH-Easy & AIME25 \\
    \midrule
    Base                & 1.39 & 1.56 & 1.53 \\
    SFT                 & 3.17 & 3.52 & 4.40 \\
    Vanilla GSPO        & 6.63 & 6.68 & 7.67 \\
    Novelty bonus       & 4.14 & 4.57 & 5.16 \\
    \bottomrule
  \end{tabular}
\end{table}

\begin{table}[h]
  \caption{Mean \textsc{hypothesize} (left) and \textsc{backtrack} (right)
    counts per trace.}
  \label{tab:breadth-hyp-btk}
  \centering
  \small
  \begin{tabular}{lccc@{\hskip 1.5em}ccc}
    \toprule
    & \multicolumn{3}{c}{\textsc{hypothesize}} & \multicolumn{3}{c}{\textsc{backtrack}} \\
    \cmidrule(lr){2-4}\cmidrule(lr){5-7}
    Checkpoint & Hard & Easy & AIME25 & Hard & Easy & AIME25 \\
    \midrule
    Base                & 0.02 & 0.02 & 0.01 & 0.00 & 0.00 & 0.01 \\
    SFT                 & 2.02 & 1.11 & 0.96 & 0.41 & 0.31 & 0.29 \\
    Vanilla GSPO        & 0.59 & 0.29 & 0.19 & 0.08 & 0.08 & 0.05 \\
    Novelty bonus       & 1.20 & 0.73 & 0.55 & 0.41 & 0.37 & 0.24 \\
    \bottomrule
  \end{tabular}
\end{table}

\begin{table}[h]
  \caption{Mean exploitation-motif counts per trace ($k = 3$:
    \textsc{check}$\to$\textsc{compute}$\to$\textsc{check}; $k = 5$:
    \textsc{check}$\to$\textsc{compute}$\to$\textsc{check}$\to$\textsc{compute}$\to$\textsc{check}).}
  \label{tab:breadth-motifs}
  \centering
  \small
  \begin{tabular}{lccc@{\hskip 1.5em}ccc}
    \toprule
    & \multicolumn{3}{c}{$k = 3$ motif} & \multicolumn{3}{c}{$k = 5$ motif} \\
    \cmidrule(lr){2-4}\cmidrule(lr){5-7}
    Checkpoint & Hard & Easy & AIME25 & Hard & Easy & AIME25 \\
    \midrule
    Base                & 0.00 & 0.01 & 0.01 & 0.00 & 0.00 & 0.00 \\
    SFT                 & 0.57 & 0.71 & 0.88 & 0.17 & 0.22 & 0.30 \\
    Vanilla GSPO        & 1.40 & 1.34 & 1.48 & 0.58 & 0.58 & 0.64 \\
    Novelty bonus       & 1.00 & 1.11 & 1.19 & 0.38 & 0.43 & 0.50 \\
    \bottomrule
  \end{tabular}
\end{table}

\paragraph{Where magnitudes shift.}
AIME25 has slightly higher chain depth at every checkpoint
($\sim$$1$ unit deeper than Hard) — competition-math problems require longer
reasoning chains than the OlymMATH-Hard distribution. Easy has slightly weaker
HYP/BTK suppression ratios under vanilla GSPO than Hard, consistent with Easy
saturating earlier ($\mathrm{SFT}~\texttt{pass@32} = 72\%$): the SFT
distribution there is already close to ceiling, leaving less room for vanilla
RL to differentiate. The AIME25 novelty-bonus cell has the smallest sample
size (960 traces; $30$ docs $\times$ $32$ rollouts) and is the most
sample-size-noisy column. Across all metrics, no benchmark contradicts the
direction of any \S\ref{sec:rl-deepens}--\S\ref{sec:rl-collapses} ranking.

\begin{figure}[h]
  \centering
  \includegraphics[width=0.7\linewidth]{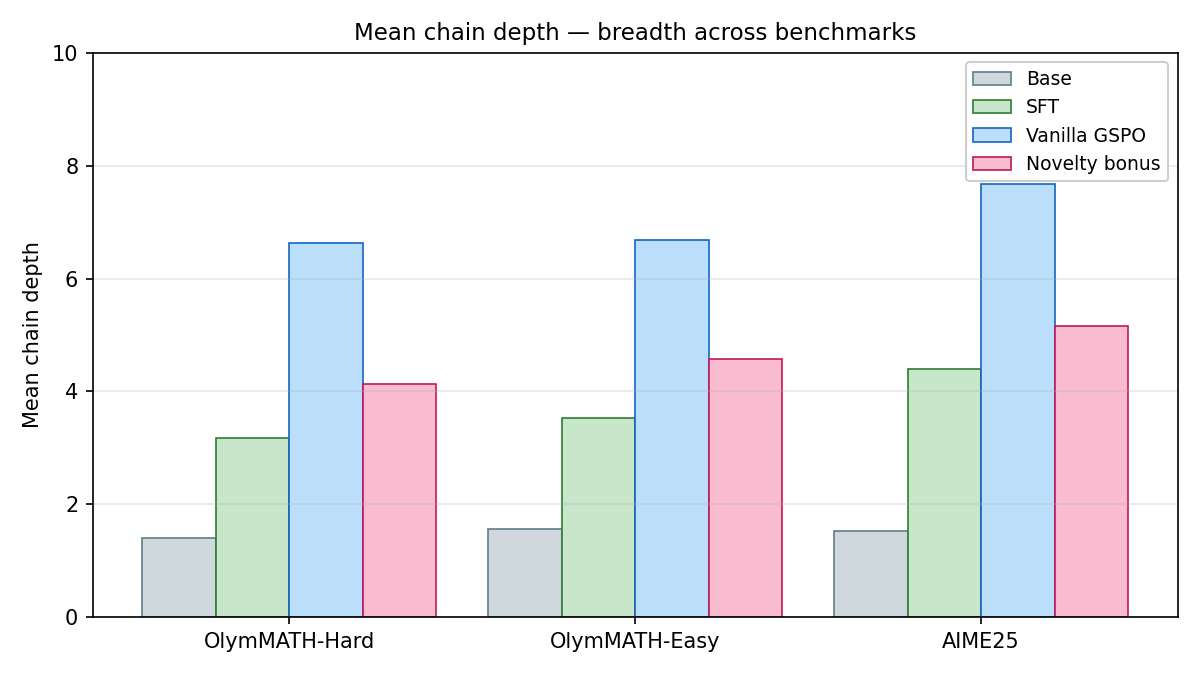}
  \caption{Mean chain depth across the four checkpoints on OlymMATH-Hard,
    OlymMATH-Easy, and AIME25.}
  \label{fig:breadth-chain-depth}
\end{figure}

\begin{figure}[h]
  \centering
  \includegraphics[width=0.85\linewidth]{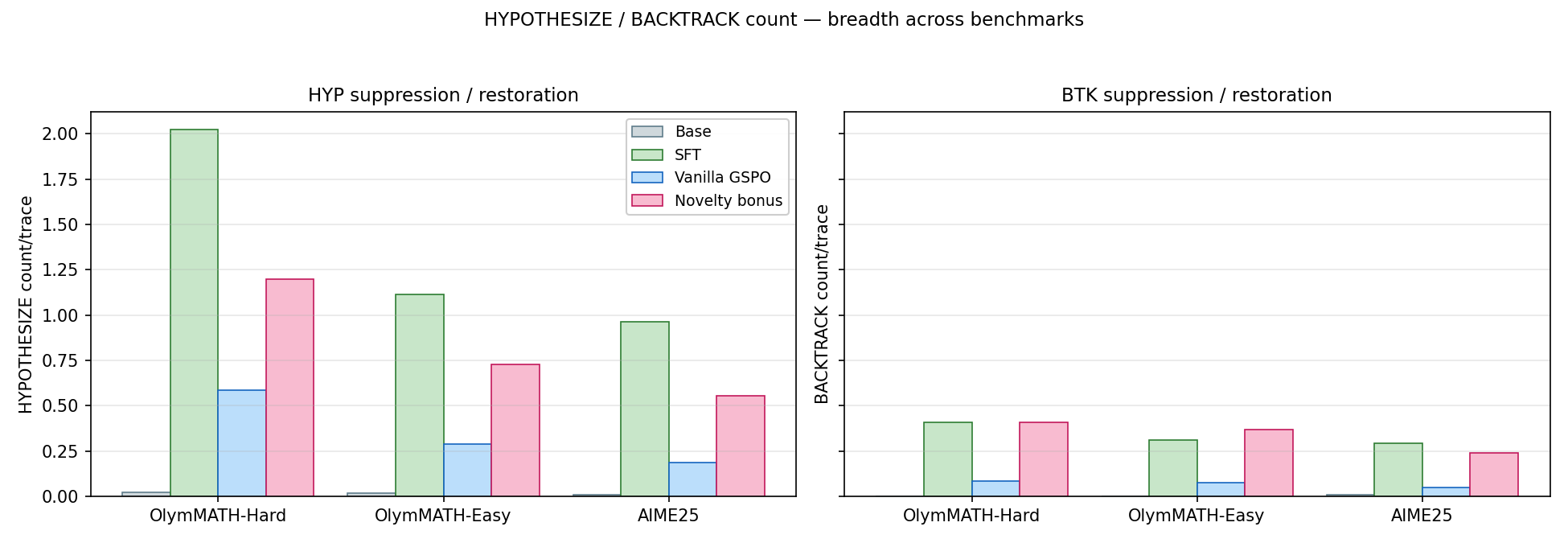}
  \caption{Mean \textsc{hypothesize} and \textsc{backtrack} counts per trace,
    faceted by benchmark.}
  \label{fig:breadth-hyp-btk}
\end{figure}

\begin{figure}[h]
  \centering
  \includegraphics[width=0.85\linewidth]{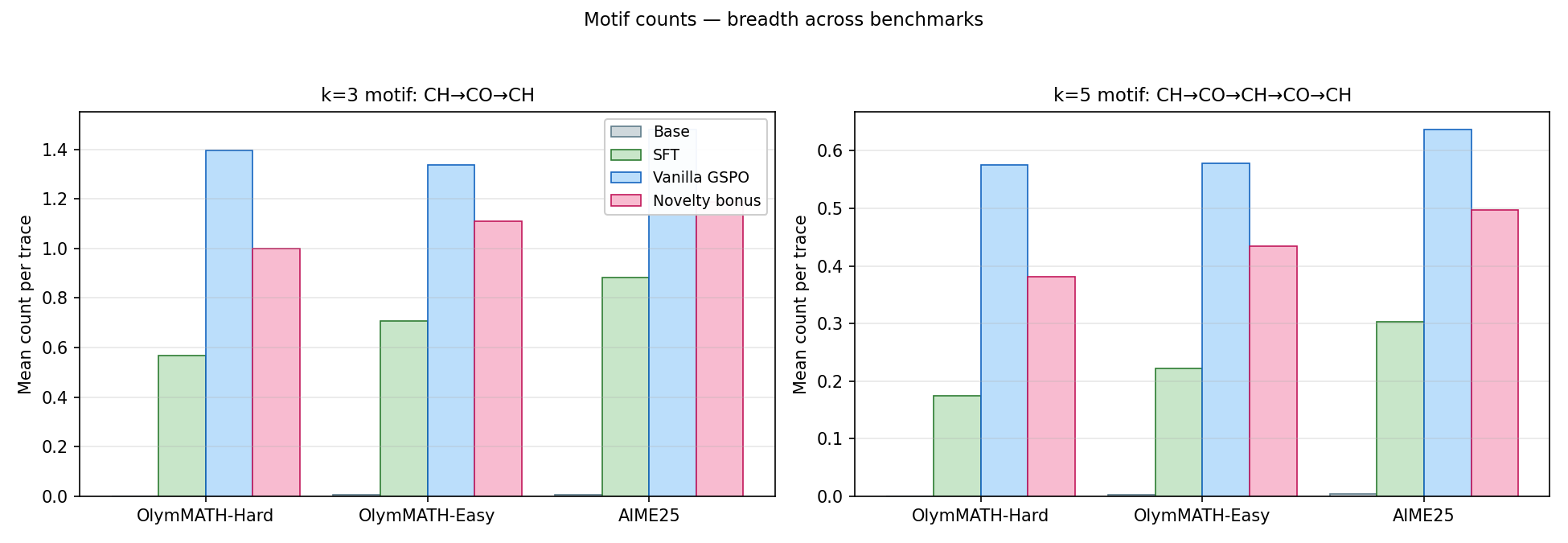}
  \caption{Mean $k = 3$ and $k = 5$ exploitation-motif counts per trace,
    faceted by benchmark.}
  \label{fig:breadth-motifs}
\end{figure}

\section{Depth shift on puzzles and within-checkpoint scaling}
\label{app:within-checkpoint-depth}

Section~\ref{sec:rl-deepens} reports that vanilla RL roughly doubles chain
depth on OlymMATH-Hard. We verify here that (i) the same depth shift is
present on harder OOD puzzles---so the math depth shift is not a
math-specific artefact---and (ii) within-checkpoint, depth scales with
task complexity, ruling out a mechanical ``bigger-input-longer-output''
reading.

\begin{figure}[h]
  \centering
  \begin{minipage}{0.42\linewidth}
    \centering
    \includegraphics[width=\linewidth]{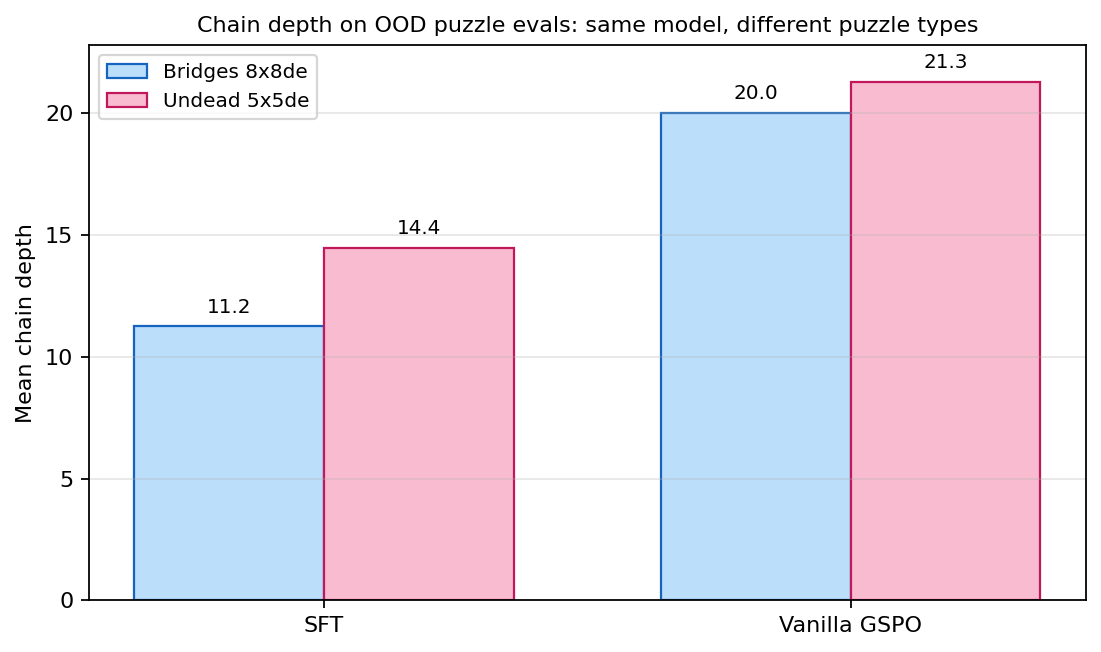}
    \subcaption{Same model, different puzzle type.}
    \label{fig:chain-depth-per-puzzle}
  \end{minipage}\hfill
  \begin{minipage}{0.56\linewidth}
    \centering
    \includegraphics[width=\linewidth]{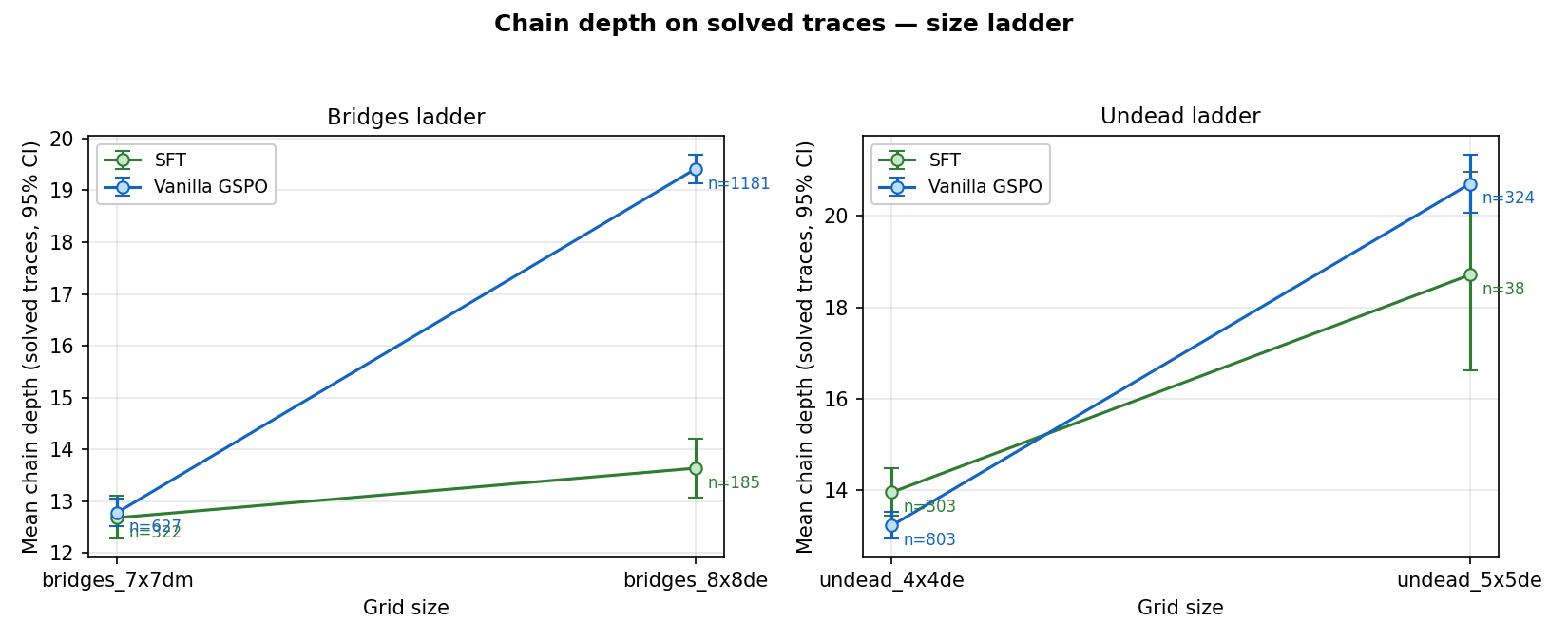}
    \subcaption{Same model and puzzle type, different grid size (solved
      traces only; $95\%$ CI).}
    \label{fig:chain-depth-size-ladder}
  \end{minipage}
  \caption{Within-checkpoint depth scaling. Chain depth tracks task
    complexity at inference: \textbf{(left)} undead is deeper than bridges
    despite a smaller grid; \textbf{(right)} a one-step grid increase lifts
    chain depth far more under vanilla GSPO than under SFT. Vanilla GSPO is
    a more aggressive depth-allocator.}
  \label{fig:chain-depth-allocation}
\end{figure}

\paragraph{The SFT$\to$vanilla GSPO depth shift appears on OOD puzzles too.}
On \emph{solved} traces at the largest puzzle grids in our test suite
(held out from training, which used smaller grids), mean chain depth rises
substantially from SFT to vanilla GSPO. On Bridges $8{\times}8$, depth goes
from $13.6$ [$13.1$, $14.2$] at SFT ($n_\text{solved}{=}185$) to $19.4$
[$19.1$, $19.7$] at vanilla GSPO ($n_\text{solved}{=}1181$)---a $+5.8$
shift with non-overlapping $95\%$ bootstrap CIs. On Undead $5{\times}5$,
depth rises from $18.7$ [$16.6$, $21.0$] (SFT, $n{=}38$) to $20.7$ [$20.1$,
$21.3$] (vanilla GSPO, $n{=}324$); the SFT estimate is noisy due to low
solved counts but the direction matches Bridges. The depth allocation
learned during puzzle RL surfaces in both held-out math and harder puzzle
generalizations (Figure~\ref{fig:chain-depth-size-ladder}).

\paragraph{Within-checkpoint, depth scales with task complexity.}
Holding the model fixed and varying the puzzle type
(Figure~\ref{fig:chain-depth-per-puzzle}), Undead $5{\times}5$ elicits
deeper chains than Bridges $8{\times}8$ in both SFT ($14.4$ vs $11.2$) and
vanilla GSPO ($21.3$ vs $20.0$, averaged over all traces)---despite the smaller grid, ruling out the
mechanical ``bigger grid $\Rightarrow$ longer trace'' story. We attribute
this to the complexity of line-tracing under reflection in undead vs.\ the
simpler bridge-count constraint.

Holding the model and puzzle type fixed and varying only the grid size on
\emph{solved} traces (Figure~\ref{fig:chain-depth-size-ladder}, $95\%$
bootstrap CIs), Bridges $7{\times}7 \to 8{\times}8$ raises mean chain depth
by $+0.96$ for SFT but by $+6.63$ for vanilla GSPO ($6.9\times$ steeper
slope); on Undead $4{\times}4 \to 5{\times}5$ the slopes are $+4.75$ vs
$+7.47$. Vanilla GSPO is a \emph{more aggressive depth-allocator} than SFT:
when the task demands more reasoning, it stretches further. This is
consistent with the RL stage training on harder puzzles than SFT, learning
to compose longer compute--check chains under verifiable reward.

\section{Solved-vs-unsolved chain-depth significance}
\label{app:chain-depth-sig}

Table~\ref{tab:chain-depth-sig} reports the per-checkpoint Mann--Whitney $U$
test for the solved-vs-unsolved chain-depth split shown in
Figure~\ref{fig:confirmation-evidence}(b). The figure itself plots only the
two means and the n-of-solved annotation; significance details are kept here
for reference.

\begin{table}[h]
  \caption{OlymMATH-Hard chain depth, solved vs unsolved traces, per checkpoint.
    Mann--Whitney $U$ tests are two-sided and unadjusted; sample sizes are
    $n_{\text{solved}}$ and $n_{\text{unsolved}}$ respectively, summing to the
    pass@$32$ trace pool (3{,}200 per checkpoint).}
  \label{tab:chain-depth-sig}
  \centering
  \small
  \begin{tabular}{lcccccc}
    \toprule
    Checkpoint & $\bar d_{\text{solved}}$ & $\bar d_{\text{unsolved}}$ & $\Delta$ & $U$ & $p$ & $n_{\text{solved}}$ \\
    \midrule
    Base          & --      & 0.83 & --      & --      & --      & 0 \\
    SFT           & 7.35    & 6.06 & $+1.29$ & $159{,}073$ & $0.003$ & 86 \\
    Vanilla GSPO  & 9.96    & 14.46 & $-4.50$ & $121{,}328$ & $0.020$ & 91 \\
    Novelty bonus & 10.29   & 8.27  & $+2.02$ & $184{,}836$ & $0.009$ & 104 \\
    \bottomrule
  \end{tabular}
\end{table}

\section{Training hyperparameters}
\label{app:hyperparams}

All experiments use \texttt{allenai/OLMo-3-7B-Instruct-SFT} as the base. Both stages
train LoRA adapters (rank $64$, \texttt{all-linear}); stage-specific $\alpha$ values are
in Tables~\ref{tab:hyper-sft}--\ref{tab:hyper-rl}.

\begin{table}[h]
  \caption{Stage~1 (SFT) hyperparameters. The SFT corpus contains rejection-sampled
    DSR puzzle traces only; we keep traces whose final answer verifies against the
    puzzle scorer and discard the rest. Epoch~5 is selected as the SFT endpoint (best
    held-out puzzle pass@1).}
  \label{tab:hyper-sft}
  \centering
  \small
  \begin{tabular}{ll}
    \toprule
    Setting & Value \\
    \midrule
    Base model              & OLMo-3-7B-Instruct-SFT \\
    LoRA rank / $\alpha$ / target  & $64$ / $64$ / \texttt{all-linear} \\
    Optimizer / learning rate     & AdamW / $2{\times}10^{-4}$ \\
    Train batch size              & $32$ (8 GPUs $\times$ micro-batch $1$ $\times$ $4$ gradient-accum.\ steps) \\
    Max sequence length           & $32{,}000$ (corpus filtered to $\le$ $28$k tokens) \\
    Epochs / endpoint             & $8$ trained / epoch $5$ used downstream \\
    Precision                     & bfloat16 training; fp32 LoRA merge \\
    \bottomrule
  \end{tabular}
\end{table}

\begin{table}[h]
  \caption{Stage~2 (RL) hyperparameters. The vanilla-GSPO baseline and the
    novelty-bonus variant share all of these settings; the only difference is whether
    the novelty bonus (Section~\ref{sec:novelty}, $\alpha{=}0.1$, top-$k{=}100$,
    $z$-clip $2$) is added on top of the GSPO reward.}
  \label{tab:hyper-rl}
  \centering
  \small
  \begin{tabular}{ll}
    \toprule
    Setting & Value \\
    \midrule
    Init / KL reference          & SFT epoch~5, fp32-merged checkpoint \\
    LoRA rank / $\alpha$ / target & $64$ / $128$ / \texttt{all-linear} \\
    Optimizer / learning rate    & AdamW / $5{\times}10^{-5}$ (cosine, $\min$ ratio $0.3$) \\
    Train batch size             & $128$ prompts \\
    Rollouts per prompt          & $8$ \\
    PPO mini-batch / epochs      & $1024$ / $4$ \\
    Sampling                     & $T{=}0.8$, top-$p{=}1.0$ \\
    Max prompt / response length & $3{,}000$ / $28{,}000$ tokens \\
    GSPO clip ratios             & $\epsilon_{\text{low}}{=}2.5{\times}10^{-3}$,
                                   $\epsilon_{\text{high}}{=}5{\times}10^{-3}$ \\
    KL coefficient $\beta$       & $10^{-3}$ (low-variance KL) \\
    Reward                       & exact + completion reward ($p{=}3$, changed-cell weight $2$) + $0.1\cdot$format reward \\
    Epochs over data / total steps & $1$ (no data recycling) / $70$ steps \\
    GPUs                         & $8\times$B200 \\
    \bottomrule
  \end{tabular}
\end{table}

Both runs use $n_{\text{gen}}{=}8$ on $8\times$B200 GPUs. All other RL hyperparameters are identical.

\paragraph{Verifier reward components.}
The exact term is binary: it is $1$ iff the extracted \texttt{<answer>} verifies
against the puzzle solution and $0$ otherwise. The completion reward is a dense
board-completion score used before exact solving is common: after strict shape
checks, it computes the fraction of grid cells matching the ground-truth
solution, gives cells that differ from the initial puzzle board weight $2$, and
raises the weighted fraction to power $p=3$. The format reward is an
XML-structure reward in $[0,1]$: it gives $0.25$ each for exactly one closed
\texttt{<reasoning>} tag, exactly one closed \texttt{<answer>} tag, reasoning
appearing before answer, and no duplicate opening tags. Thus the table's
$0.1\cdot$format-reward contribution is at most $0.1$ and is intended only to
stabilize response formatting.

\section{Motif extraction details}
\label{app:motif-details}

\paragraph{Span segmentation algorithm.}
Each reasoning trace is processed as follows. First, the reasoning block is extracted from
\texttt{\textless{}reasoning\textgreater{}} tags if present; otherwise the full response is used.
The text is then split at paragraph boundaries (double newlines) and further split within each
paragraph at discourse markers detected by a multi-line, case-insensitive regex covering:
numbered steps (\textit{Case N}, \textit{Step N}, \textit{Option N});
sequential connectives (\textit{First}, \textit{Second}, \textit{Then});
hypothetical openers (\textit{Suppose}, \textit{Assume}, \textit{Let's});
verification openers (\textit{Check}, \textit{Verify}, \textit{Confirm});
self-correction markers (\textit{Wait}, \textit{Actually}, \textit{Hmm});
redirection markers (\textit{Instead}, \textit{Alternatively}, \textit{Going back});
conclusion markers (\textit{Therefore}, \textit{Hence}, \textit{Thus});
and failure markers (\textit{Contradiction}, \textit{This doesn't work}).
Short fragments below $80$ tokens are merged into an accumulation buffer until the threshold
is reached; long fragments above $250$ tokens are split at sentence boundaries (period,
exclamation mark, or question mark followed by whitespace). A trailing buffer below $40$
tokens is merged into the previous span rather than emitted as a standalone fragment.
Token counts during merging use a character-based estimate (character count divided by~$4$)
for speed; the OLMo3 tokenizer is invoked only for the final token count on each emitted span.

\paragraph{Span classifier training.}
We fine-tune Qwen3-1.7B~\cite{qwen3technicalreport} as a nine-way sequence classifier over the primitive vocabulary of Table~\ref{tab:primitives}. Training data consists of $35$k judge-labeled spans: $17$k from puzzle reasoning traces and $18$k from mathematics reasoning traces. Labels were produced by a jury of LLM judges (DeepSeek-V3.2 and DeepSeek-Chat~\cite{liu2025deepseekv32}), iterating prompt refinements until inter-judge agreement reached $86\%$ on a held-out span set. The resulting classifier reaches $0.80$ macro-F1 against judge consensus on a held-out test set. All primitive sequences in Sections~\ref{sec:sft-induces}--\ref{sec:confirmation} use this classifier.

\paragraph{Sequence representation.}
For each reasoning trace, our classifier of Section~\ref{sec:framework} produces
a sequence of primitive labels, one per detected episode. We restrict to the
$8$ reasoning labels (\textsc{plan, setup, enumerate, hypothesize, compute, check,
backtrack, summarize}) and treat \textsc{other} as a boundary token \texttt{\_\_BREAK\_\_}
that prevents adjacency across formatting/meta spans
(e.g.\ \texttt{[CHECK, OTHER, SETUP]}$\to$\texttt{[CHECK, \_\_BREAK\_\_, SETUP]}, so
\texttt{CHECK}$\to$\texttt{SETUP} is not counted as a bigram). \textsc{other} is
predominantly output formatting in our traces ($91.5\%$ of \textsc{other} spans
appear in the last $20\%$ of the trace) and treating it as a reasoning token would
inflate cross-section adjacencies.

\paragraph{$k$-gram extraction.}
For each $k \in \{2, \ldots, 15\}$ we extract all length-$k$ subsequences via a
sliding window and count motifs that do not cross a \texttt{\_\_BREAK\_\_}. We use
the observed vocabulary only (full enumeration is infeasible: $9^{10}{\approx}3.5\!\times\!10^{9}$).
Counts are length-normalised per trace as
$\mathrm{freq}(m) = \mathrm{count}(m) / (L - k + 1)$ where $L$ is the post-break
sequence length, and motifs with raw count below $5$ (for $k{\le}5$) or $10$ (for
$k{>}5$) are filtered as noise.

\paragraph{Reporting.}
Section~\ref{sec:rl-deepens} reports per-trace counts for three concrete motifs
(Figure~\ref{fig:motif-examples}). For consistency with related work that does not
distinguish overlapping vs non-overlapping matches, those counts use
non-overlapping matches; cross-checkpoint comparisons are unaffected by the choice.

\paragraph{Motif categories used in Figure~\ref{fig:primitive-motif-distributions}.}
The Section~\ref{sec:sft-induces} motif row aggregates length-$3$ trigrams into
three curated categories. Let $w = (w_1, w_2, w_3)$ be a length-$3$ window over
the primitive sequence; let $\mathrm{ALLOWED\_MIDDLE} = \{$\textsc{plan},
\textsc{setup}, \textsc{enumerate}, \textsc{compute}, \textsc{summarize},
\textsc{other}$\}$. The three category predicates are:
\begin{itemize}[leftmargin=*,topsep=2pt,itemsep=2pt]
  \item \emph{Recovery}: $\textsc{hypothesize} \in w$ \textbf{and} $\textsc{backtrack} \in w$
        (order-agnostic; matches any trigram window containing both recovery primitives).
  \item \emph{Exploitation}: $w = (\textsc{compute}, \textsc{check}, \textsc{compute})$
        (anchored on \textsc{compute}; the canonical compute$\leftrightarrow$check
        alternation chain).
  \item \emph{Verification}: $w_1 = \textsc{check}$, $w_3 = \textsc{check}$,
        $w_2 \in \mathrm{ALLOWED\_MIDDLE}$ (anchored on \textsc{check} with a non-recovery,
        non-\textsc{check} middle slot).
\end{itemize}
The middle-slot exclusions in Verification are deliberate. Excluding \textsc{check}
from $w_2$ prevents long runs of consecutive \textsc{check}s from inflating the
verification count via the sliding window without adding semantic content (a
sticky-\textsc{check} artefact rather than a check--act--check pattern). Excluding
\textsc{hypothesize} and \textsc{backtrack} from $w_2$ keeps Verification disjoint
from Recovery: a window like $(\textsc{check}, \textsc{hypothesize}, \textsc{check})$
that would otherwise satisfy both predicates is counted under Recovery only. The
Exploitation predicate is exact-match (\textsc{compute}-anchored), so it is also
disjoint from the other two by construction. This three-way disjointness is what
makes the three category bars in Figure~\ref{fig:primitive-motif-distributions}
sum interpretably without double-counting. The $k=5$ extension used elsewhere
($\textsc{check}$$\to$$x$$\to$$\textsc{check}$$\to$$y$$\to$$\textsc{check}$ with
$x, y \in \mathrm{ALLOWED\_MIDDLE}$ for verification, and the analogous
$\textsc{check}$$\to$$\textsc{compute}$$\to$$\textsc{check}$$\to$$\textsc{compute}$$\to$$\textsc{check}$ for exploitation) preserves the same anchor and
disjointness rules.

\section{Metric definitions}
\label{app:metrics}

Let $s = (s_1, \ldots, s_L)$ denote the primitive sequence of a trace
(post-\textsc{other}-filtering as in Appendix~\ref{app:motif-details}) and let
$T$ denote the trace's response token count.

\paragraph{Chain depth.}
We define an \emph{exploit run} as a maximal contiguous block of primitives drawn
from $\mathrm{EXPLOIT} = \{\textsc{compute}, \textsc{check}, \textsc{setup}\}$. The
chain depth of a trace is the length of its longest exploit run:
\begin{equation*}
  \mathrm{depth}(s) \;=\; \max_{i \le j} \big\{\, j - i + 1 \;:\; s_i, \ldots, s_j \in \mathrm{EXPLOIT} \,\big\}.
\end{equation*}
We include \textsc{setup} alongside \textsc{compute} and \textsc{check} because
it represents on-track implementation work (introducing notation, defining
auxiliary variables) rather than strategic decomposition or branching;
excluding it would shatter compute--check runs every time the model names a
new variable. The ranking is robust to this choice: with
$\mathrm{EXPLOIT} = \{\textsc{compute}, \textsc{check}\}$ (no \textsc{setup}),
mean chain depths on OlymMATH-Hard are Base $0.83$, SFT $2.27$, vanilla GSPO
$4.70$, novelty $3.25$ — the same Base $<$ SFT $<$ vanilla GSPO ordering with
novelty between SFT and vanilla GSPO, and the SFT$\to$vanilla GSPO ``roughly
doubles'' claim ($2.07\times$ vs $2.09\times$ with \textsc{setup}). We report
mean depth and the $90$-th percentile per checkpoint.

\paragraph{Mean exploit-run length.}
Let $r_1, r_2, \ldots, r_M$ be the lengths of the maximal exploit runs in $s$
(a run is a maximal contiguous block of primitives in $\mathrm{EXPLOIT}$).
The mean exploit-run length of a trace is
\begin{equation*}
  \mathrm{mean\_run}(s) \;=\; \frac{1}{M} \sum_{m=1}^{M} r_m,
\end{equation*}
undefined when $M = 0$; we report mean over traces with $M \ge 1$. Unlike chain
depth, which picks the single longest run, mean run length averages all
exploit runs in a trace and is therefore sensitive to whether exploit
primitives are clustered into a few long runs (high mean) or scattered as
many short runs (low mean). We use it alongside chain depth to ensure the
depth-of-exploitation claim is not an artefact of one long run with many
scattered short runs around it.

\section{Alternative diversity mechanisms}
\label{app:diversity-alternatives}

Production GSPO runs at our scale take roughly $70$ wall-clock hours on
$8\times$B200 GPUs per configuration ($\approx$560 GPU-hours), making it
impractical to iterate diversity-bonus designs at full setup. We therefore ran a smaller
\emph{mini} ablation setup---$2$ puzzles (Bridges and Undead) and a $12$k-token
response budget per rollout, against the $4$-puzzle, $28$k-token production
setup---roughly $6\times$ cheaper per run. We use mini to compare
diversity-bonus candidates and run only the best at production scale. Other
diversity-promoting interventions applied to the same setup either collapse
or fall short on stability (Table~\ref{tab:diversity}).

At mini scale, the loss-level entropy bonus and our novelty bonus both
reach \texttt{pass@32}\,$=28.0\%$ at their best step---a tie on peak
performance. The runs differ on training stability: the entropy-bonus run
peaks at step $5$ and degrades thereafter, while the novelty-bonus run
trains stably through step $20$. Entropy regularisation also does not target
the specific \textsc{hypothesize}/\textsc{backtrack} subspace---it pulls
the whole distribution toward uniformity. The entropy-advantage variants
of~\citet{cheng2025reasoningexploration} collapsed below baseline ($17.0\%$
no-KL, $14.0\%$ with KL).

These pilots should be read as a screening study rather than a full-scale
ranking of diversity mechanisms. The mini setup omits five puzzles and uses a
shorter response budget, and we did not train every alternative at production
scale. The evidence supports two narrower claims: first, the frozen-reference
novelty bonus was the most stable candidate in our pilot set; second, when
scaled to the production setup, it improved over the vanilla-GSPO production
baseline ($36\%$ vs $29\%$ \texttt{pass@32}). A full production-scale comparison
against entropy and trajectory-rarity baselines remains future work.

\begin{table}[h]
  \caption{Alternative diversity mechanisms on OlymMATH-Hard. All
    entries use GSPO at the same data and clip configuration; only the
    diversity mechanism differs. Numbers are best-step \texttt{pass@8} /
    \texttt{pass@32} over $n = 32$ rollouts.}
  \label{tab:diversity}
  \centering
  \small
  \begin{tabular}{llcc}
    \toprule
    Method & Scale & \texttt{pass@8} & \texttt{pass@32} \\
    \midrule
    GSPO baseline                                                        & production & 12.6 & 29.0 \\
    Loss-level entropy bonus (unstable)                                  & mini       & 12.4 & 28.0 \\
    Entropy-advantage no-KL~\cite{cheng2025reasoningexploration}         & mini       & 8.9  & 17.0 \\
    Entropy-advantage +KL                                                & mini       & 5.5  & 14.0 \\
    Novelty bonus                                                        & mini       & 13.1 & 28.0 \\
    \textbf{Novelty bonus (ours)}                                        & \textbf{production} & \textbf{16.4} & \textbf{36.0} \\
    \bottomrule
  \end{tabular}
\end{table}

\section{Novelty-bonus algorithm}
\label{app:novelty-pseudocode}

Figure~\ref{alg:novelty} gives the per-step computation that augments the GSPO
reward with the frozen-reference novelty bonus described in
Section~\ref{sec:novelty}. The bonus is reward shaping: it is computed once per
rollout under the \emph{frozen} SFT reference policy $\pi_{\text{ref}}$ (no
gradients), then added to the last valid token's reward before advantage
computation. All other GSPO machinery (clip ratios, KL penalty, optimisation)
is unchanged.

\begin{figure}[h]
  \centering
  \fbox{\parbox{0.95\linewidth}{\small
  \textbf{Top-$k$ frozen-reference novelty bonus (per training step).}\\[2pt]
  \textbf{Inputs:} rollouts $\{r_i\}_{i=1}^{B}$ grouped by prompt id $\mathrm{uid}(r_i)$;
  per-token frozen reference log-probs $\log \pi_{\text{ref}}(r_i)$; response masks
  $m_i$; correctness flags $a_i \in \{0,1\}$; per-token rewards $R_i$;
  hyper-parameters $\alpha{=}0.1$, $k{=}100$, $z_{\text{clip}}{=}2$.
  \begin{enumerate}[leftmargin=*,itemsep=2pt,topsep=4pt]
  \item For each rollout $i$, compute the masked per-token NLL
    $\mathrm{neg\_lp}_i = -\log \pi_{\text{ref}}(r_i) \odot m_i$ and the top-$k$
    score
    $s_i = \tfrac{1}{k}\!\sum_{t \in \mathrm{TopK}_k(\mathrm{neg\_lp}_i)} \mathrm{neg\_lp}_{i,t}$.
  \item For each prompt group $g$, let $C_g = \{ i \in g : a_i = 1 \}$ be the
    correct rollouts. Skip the group if $|C_g| < 2$ or $\mathrm{std}(\{s_i\}_{i\in C_g}) < 10^{-6}$.
  \item Compute within-group statistics
    $\mu_g, \sigma_g = \mathrm{mean}, \mathrm{std}$ of $\{s_i\}_{i \in C_g}$, and
    set
    $z_i = \mathrm{clip}((s_i - \mu_g)/\sigma_g,\, -z_{\text{clip}},\, +z_{\text{clip}})$
    for each $i \in C_g$.
  \item Add the bonus at the last valid token of each correct rollout:
    $R_i[\text{last valid}] \mathrel{+}= \alpha \cdot z_i$.
  \item Compute GSPO advantages from the modified token-level rewards $R$ as usual.
  \end{enumerate}
  }}
  \caption{Pseudocode for the frozen-reference top-$k$ novelty bonus described in
    Section~\ref{sec:novelty}. The bonus is reward shaping: it is computed once per
    rollout under the frozen SFT reference and added to the last valid token's
    reward before advantage computation. All other GSPO machinery (clip ratios, KL
    penalty, optimisation) is unchanged.}
  \label{alg:novelty}
\end{figure}

The bonus is applied only to correct rollouts ($a_i = 1$): incorrect rollouts
already receive zero base reward, so adding novelty there would amount to
rewarding incorrect novel reasoning. Within-group $z$-scoring with the
$z_{\text{clip}}{=}2$ cap bounds $|\alpha z| \le 0.2$, i.e.\ at most $20\%$ of the
maximum base reward, which keeps the bonus shaping rather than dominant.

\section{Within-group signal analysis}
\label{app:novelty-signal-analysis}

The novelty bonus only has signal if SFT perplexity varies meaningfully across
correct rollouts of the \emph{same} prompt: with no within-group spread, the
$z$-score in Figure~\ref{alg:novelty} is undefined and the bonus collapses to
zero. We measured this directly on $100$ puzzle prompts ($32$ rollouts each, post-SFT)
and on $100$ OlymMATH-Hard prompts under both SFT and vanilla-GSPO checkpoints.
Table~\ref{tab:novelty-signal} reports the mean within-prompt standard deviation of
two candidate per-rollout signals: (i) mean per-token negative log-probability over
the full response, and (ii) mean negative log-probability of the top-$k$ most
surprising tokens.

\begin{table}[h]
  \caption{Within-prompt standard deviation (nats/token) of per-rollout SFT-NLL
    signals, averaged over $100$ prompts. Mean NLL has near-zero spread: at our
    response lengths ($\sim$$25$k tokens), per-token NLL is dominated by formatting
    and connective tokens that share probability across all reasoning paths.
    Restricting to the top-$k$ most surprising tokens isolates decision-critical
    positions and amplifies the within-group signal $10$--$15\times$.}
  \label{tab:novelty-signal}
  \centering
  \small
  \begin{tabular}{lccc}
    \toprule
    Signal & SFT puzzles & SFT math & Vanilla GSPO math \\
    \midrule
    Mean NLL (full response) & $0.024$ & $0.024$ & $0.024$ \\
    Top-$10$  NLL            & $0.18$  & $0.16$  & $0.16$  \\
    Top-$100$ NLL            & $0.276$ & $0.260$ & $0.245$ \\
    Top-$200$ NLL            & $0.27$  & $0.26$  & $0.24$  \\
    \bottomrule
  \end{tabular}
\end{table}

At a $z$-score cap of $2$ and $\alpha{=}0.1$, the mean-NLL signal would yield bonus
magnitudes around $\pm 2 \times 0.024 \times 0.1 \approx 0.005$ on a base reward of
order $1$---below noise. The top-$100$ signal yields $\approx 0.05$, which is
where the bonus becomes a meaningful fraction of the base reward. We chose
$k{=}100$ because the within-group std saturates between $k{=}100$ and $k{=}200$
(no further amplification), and smaller $k$ becomes noisier across prompts of
varying length.

\section{Per-problem recovery split on OlymMATH-Hard}
\label{app:per-problem-recovery-split}

Section~\ref{sec:confirmation} focuses on the two diagnostic subsets for the
selectivity claim: problems solved by both vanilla GSPO and the novelty-bonus
variant, and problems solved only by the novelty-bonus variant. The full
per-problem accounting over the $100$ OlymMATH-Hard items is:

\begin{table}[h]
  \caption{Per-problem solve-set split used in
    Section~\ref{sec:confirmation}. A problem is counted as solved by a
    checkpoint if at least one of its $32$ rollouts verifies.}
  \label{tab:confirmation-problem-split}
  \centering
  \small
  \begin{tabular}{lc}
    \toprule
    Outcome bucket & Number of problems \\
    \midrule
    Solved by both vanilla GSPO and novelty bonus & $20$ \\
    Solved only by novelty bonus                  & $16$ \\
    Solved only by vanilla GSPO                   & $9$ \\
    Solved by neither                             & $55$ \\
    \bottomrule
  \end{tabular}
\end{table}

The primitive-count split shows the same selectivity pattern as the motif
analysis in the body. On the shared problems, vanilla GSPO uses recovery
primitives at a low rate (HYP $0.63$, BTK $0.08$ per trace), and the novelty
bonus is higher but still modest (HYP $1.34$, BTK $0.38$). On
unique-to-novelty problems, vanilla GSPO remains flat (HYP $0.63$, BTK $0.10$),
while the novelty bonus rises (HYP $1.51$, BTK $0.43$); restricting to the
actually solved novelty-bonus traces on these problems raises the rates further
(HYP $2.25$, BTK $0.46$). Thus the recovered primitives are not deployed
uniformly across all problems: they concentrate most strongly on the solved
traces where vanilla GSPO fails.

\section{Paired CoT inspection on OlymMATH-Hard}
\label{app:cot-inspection}

This appendix reproduces four worked OlymMATH-Hard rollouts to illustrate the
mechanisms claimed quantitatively in the body. \emph{Part~A}
(\S\ref{sec:rl-deepens}) pairs an SFT rollout against a vanilla-GSPO rollout
where both are correct, to show that vanilla GSPO improves execution without
changing the set of approaches. \emph{Part~B} (\S\ref{sec:confirmation}) pairs
a vanilla-GSPO rollout that fails against a novelty-bonus rollout that
succeeds, on problems vanilla GSPO never solves, to show that the novelty bonus's gain
comes from in-trace recovery rather than from new mathematical machinery.
Throughout, we score every rollout with \texttt{math\_verify}, consistent with
the scoring used across all main-paper evaluations.

\subsection*{Part A: Depth of execution (\S\ref{sec:rl-deepens})}

We pair, for the same problem, an SFT rollout and a vanilla-GSPO rollout that
are both correct and share the same opening primitives. We pick two problems
whose chain depths fall near each checkpoint's mean ($\bar d \approx 3.2$ for
SFT, $\bar d \approx 6.6$ for vanilla GSPO).

\begin{figure}[h]
  \centering
  \includegraphics[width=\linewidth]{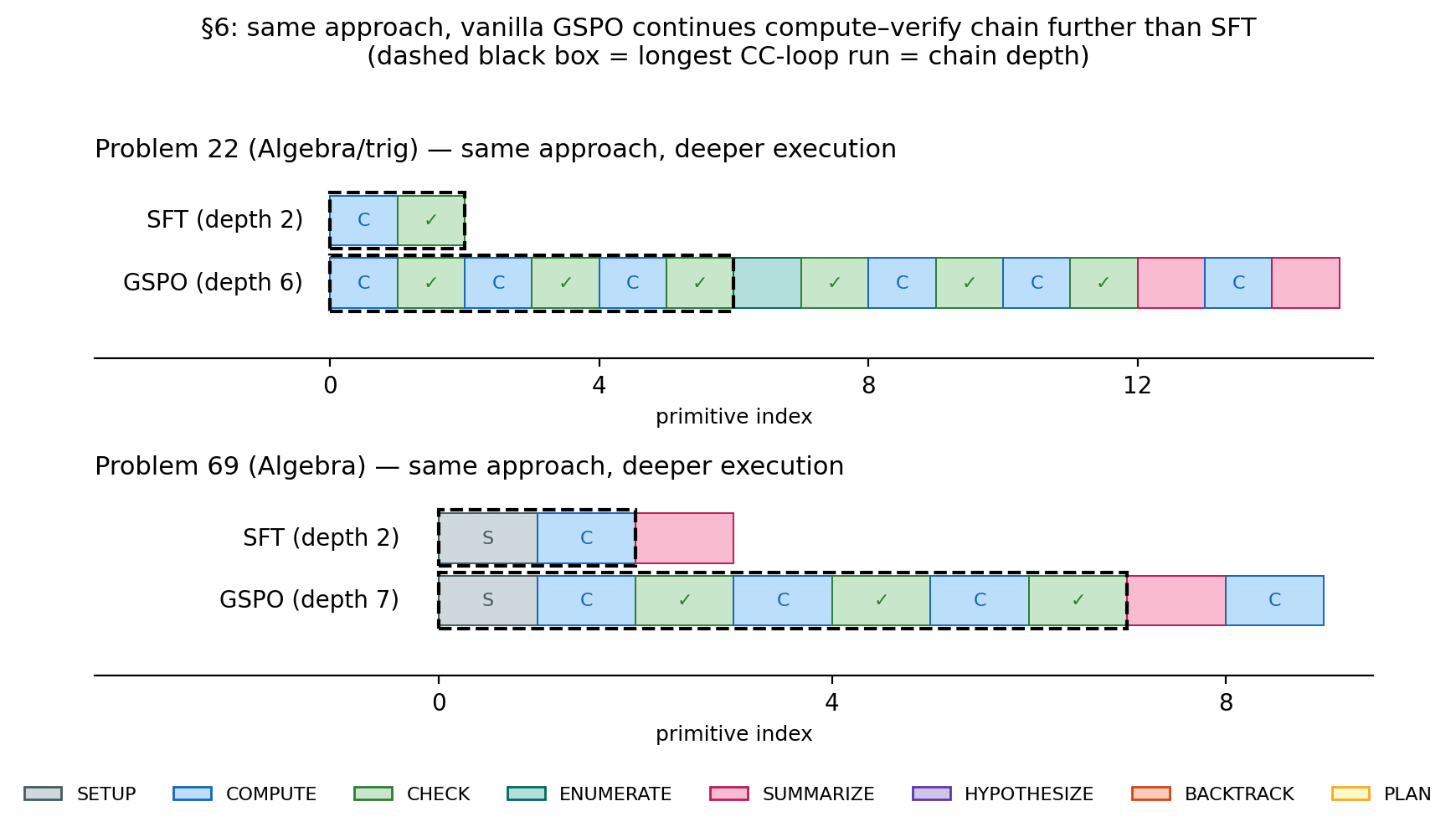}
  \caption{Primitive-sequence strips for the SFT vs vanilla-GSPO pairs on
    problems $22$ and $69$. Shared opening primitives sit at indices $0$--$1$
    in both pairs; the longest exploit run (dashed black box) is the
    chain-depth metric reported in \S\ref{sec:rl-deepens}.}
  \label{fig:cot-inspection-depth}
\end{figure}

In both cases the SFT and GSPO rollouts converge on the same algebraic
reformulation in the same primitive sequence: single-variable substitution
yielding a quadratic, then a vertex/endpoint range argument. They differ in
how much compute--verify alternation follows: GSPO continues for one to three
additional \textsc{compute}$\to$\textsc{check} cycles---numerical spot checks,
boundary tests, sign discussion---before stopping, while SFT terminates after
one verification pass. No new approach appears in the GSPO trace that is
absent from the SFT trace. Figure~\ref{fig:cot-inspection-depth} shows the
primitive-sequence strips for both pairs; the longest \textsc{compute}/\textsc{check}
run (chain depth) is annotated.

\paragraph{Problem 1 (Algebra, trigonometric optimisation).}
Let $f(x) = \sin^4 \omega x - \sin \omega x \cos \omega x + \cos^4 \omega x$
with $\omega > 0$. If there exist $a, b \in [0, \pi]$ with
$f(a) + f(b) = 9/4$, find the minimum $\omega$. Gold answer: $\omega = 7/12$.
SFT trace depth $2$; vanilla-GSPO trace depth $6$.

\begin{quote}
\small\textbf{SFT (chain depth 2):}\\
\texttt{[COMPUTE]} Use $\sin^4\theta + \cos^4\theta = 1 - 2\sin^2\theta\cos^2\theta$.
Let $t = \sin\omega x \cos\omega x$, so $f(x) = 1 - 2t^2 - t$. With
$t = \tfrac12 \sin(2\omega x)$ and $y = \sin(2\omega x) \in [-1, 1]$ rewrite
$f(y) = 1 - \tfrac12 y^2 - \tfrac12 y$. Critical point at $y = -\tfrac12$,
giving $f = 11/8$; endpoints $f(\pm 1) = 0$.\\
\texttt{[CHECK]} So $\max f = 11/8$ and $2 \cdot 11/8 = 11/4 \ge 9/4$, hence
$f(a)+f(b)=9/4$ is achievable. Need $\sin 2\omega x = -\tfrac12$,
i.e.\ $2\omega x \in \{7\pi/6, 11\pi/6\} \pmod{2\pi}$. For $x \in [0, \pi]$
this requires $2\omega\pi \ge 7\pi/6$, hence $\omega \ge 7/12$. Minimum
$\boxed{7/12}$.
\end{quote}

\begin{quote}
\small\textbf{Vanilla GSPO (chain depth 6):}\\
\texttt{[COMPUTE]} Same substitution $t = \sin\omega x \cos\omega x$; same
quadratic $g(t) = -2t^2 - t + 1$. Vertex at $t = -1/4$, $g(-1/4) = 9/8$;
endpoints $g(\pm 1/2) \in \{0, 1\}$. Range of $g$ on $[-1/2, 1/2]$ is
$[0, 9/8]$.\\
\texttt{[CHECK]} To hit $f(a)+f(b)=9/4$ the only option is
$g(t_a) = g(t_b) = 9/8$, hence $t_a = t_b = -1/4$. Translating:
$\sin 2\omega a = \sin 2\omega b = -\tfrac12$, with $a, b \in [0, \pi]$ giving
$\omega \ge 7/12$.\\
\texttt{[COMPUTE]} Solve $a = 7\pi/(12\omega)$ and $a = 11\pi/(12\omega)$ in
$[0, \pi]$; the $11\pi/6$ solution forces $\omega \ge 11/12$, the $7\pi/6$
solution only $\omega \ge 7/12$. Tentative minimum $\omega = 11/12$ [\ldots]\\
\texttt{[CHECK]} Re-derive at $b = \pi$, $\omega = 11/12$: first attempt gives
$t_b = 1/4$ (``Wait, this contradicts earlier calculation. Let me recheck.'').
Recompute: $\sin(2\omega b) = \sin(11\pi/6) = -\sin(5\pi/6) = -\tfrac12$, so
$t_b = -1/4$ after all, $f(b) = 9/8$.\\
\texttt{[COMPUTE]} Re-examine the smaller bound $\omega = 7/12$: at this
$\omega$, $a = b = \pi$ gives $2\omega a = 7\pi/6$, $\sin = -\tfrac12$,
$t = -1/4$, $f(\pi) = 9/8$, so $f(a)+f(b) = 9/4$. The pair is allowed because
the problem only asks for existence of $a, b \in [0, \pi]$.\\
\texttt{[CHECK / ENUMERATE / CHECK / COMPUTE / CHECK / SUMMARIZE]}
Test $\omega = 1/2$ (below $7/12$): no $x \in [0, \pi]$ satisfies
$\sin 2\omega x = -\tfrac12$, confirming the bound is tight from below. Conclude
$\boxed{7/12}$.
\end{quote}

Both checkpoints reduce $f$ to a single-variable quadratic and identify the
vertex condition $\sin 2\omega x = -\tfrac12$. SFT stops after one CHECK pass
and asserts the answer; GSPO continues with two additional
\textsc{compute}$\to$\textsc{check} cycles---a numerical self-correction at
$b = \pi$ and an explicit lower-bound test at $\omega = 1/2$---before
committing to $\omega = 7/12$.

\paragraph{Problem 2 (Algebra, range of a rational expression).}
If real $x, y$ satisfy $x^2 - y^2 = 4$, find the range of
$\dfrac{1}{x^2} - \dfrac{y}{x}$. Gold answer: $(-1, 1)$. SFT trace depth $2$;
vanilla-GSPO trace depth $7$.

\begin{quote}
\small\textbf{SFT (chain depth 2):}\\
\texttt{[SETUP]} $x^2 - y^2 = 4 \Rightarrow x \ne 0$. Let $y = kx$;
$x^2(1 - k^2) = 4 \Rightarrow x^2 = 4/(1-k^2)$, requiring $|k| < 1$.\\
\texttt{[COMPUTE]} $E = 1/x^2 - y/x = (1-k^2)/4 - k = -(k^2 + 4k - 1)/4$. Let
$f(k) = k^2 + 4k - 1$. Vertex at $k = -2$ (outside $(-1,1)$); $f(-1) = -4$,
$f(1) = 4$, so on $(-1, 1)$ $f$ ranges over $(-4, 4)$.\\
\texttt{[SUMMARIZE]} Hence $E = -f(k)/4$ has range $(-1, 1)$. (Earlier
``$\pm 1.25$'' boundary computations dismissed in one sentence as
miscalculated.) $\boxed{(-1, 1)}$.
\end{quote}

\begin{quote}
\small\textbf{Vanilla GSPO (chain depth 7):}\\
\texttt{[SETUP]} Note $x^2 - y^2 = (x-y)(x+y) = 4$; rewrite target as
$(1 - xy)/x^3$. Let $t = y/x \Rightarrow x^2 = 4/(1-t^2)$, forcing $|t| < 1$.\\
\texttt{[COMPUTE]} Target $= (1-t^2)/4 - t = (1 - 4t - t^2)/4 := f(t)$.\\
\texttt{[CHECK]} $f'(t) = -t/2 - 1$; $f' = 0$ at $t = -2$, outside $(-1,1)$,
so $f$ is monotonic on $(-1,1)$. Limits: $f(t \to 1^-) = -1$,
$f(t \to -1^+) = 1$.\\
\texttt{[COMPUTE]} $f'(0) = -1$, so $f$ is strictly decreasing on $(-1,1)$;
image is $(-1, 1)$ open. Endpoints not attained (would require
$x^2 - y^2 = 0$).\\
\texttt{[CHECK]} Sign of $x$ doesn't matter (both signs give same $f(t)$);
$x = 0$ impossible.\\
\texttt{[COMPUTE]} Spot-check $t = 0$ ($x = \pm 2$, target $= 1/4$);
$t = 0.5$ (target $-0.3125$, matches $f(0.5)$); $t = -0.5$ (target $0.6875$,
matches $f(-0.5)$).\\
\texttt{[CHECK]} Limit checks: $t = 0.99 \Rightarrow f \approx -0.985$
(approaches $-1$); $t = -0.99 \Rightarrow f \approx 1.245$. Reconcile: the
supremum is approached only as $t \to -1^+$ and never attained; numeric at
$t = -0.99$ is consistent with the decreasing-to-$1$ picture.\\
\texttt{[SUMMARIZE / COMPUTE]} Range $\boxed{(-1, 1)}$.
\end{quote}

Both checkpoints adopt the substitution $t = y/x$, derive the same quadratic
$f(t) = (1 - 4t - t^2)/4$ on $(-1, 1)$, and reach $(-1, 1)$. SFT terminates
after one range argument and waves off a sketchy boundary computation as a
miscalculation; GSPO continues with two additional
\textsc{compute}$\to$\textsc{check} cycles---numerical spot-checks at
$t \in \{0, \pm 0.5\}$ followed by limit checks at $t = \pm 0.99$---before
committing.

\paragraph{Selection (Part~A).}
Using \texttt{math\_verify}, $13$ OlymMATH-Hard problems
have at least one correct rollout in both SFT ($81/3200$ traces) and
vanilla GSPO ($89/3200$). Restricting to pairs that share the first two
opening primitives and place the SFT depth in $[2, 4]$ and the GSPO depth in
$[5, 9]$ yields $19$ candidate pairs across $4$ problems. The two reproduced
above were chosen for cleanest opening match, depths near each checkpoint's
mean, and algebraic-manipulation flavour, the dominant subject in the
intersection. A small number-theory and combinatorics options exist (problems
$11$, $17$, $25$, $26$, $91$) but none paired typical-depth SFT and GSPO
traces under the same-opening constraint, so they are not informative for the
$\S\ref{sec:rl-deepens}$ depth-of-exploitation claim.

\subsection*{Part B: Productive recovery (\S\ref{sec:confirmation})}

\begin{figure}[h]
  \centering
  \includegraphics[width=\linewidth]{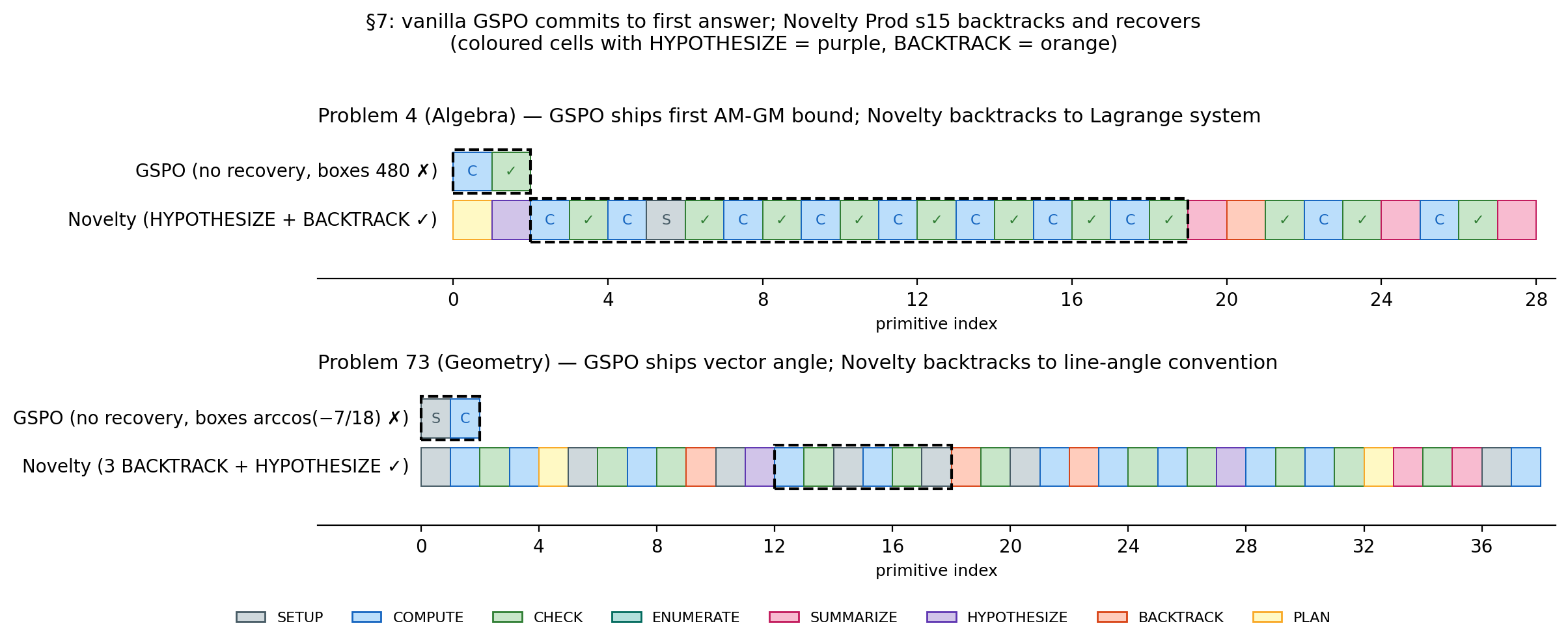}
  \caption{Primitive-sequence strips for the vanilla-GSPO vs novelty-bonus
    pairs on problems $4$ and $73$. The \textsc{hypothesize} (purple) and
    \textsc{backtrack} (orange) cells in the novelty rows mark the recovery
    pivots from the wrong first answer to the correct one; the GSPO rows
    have none.}
  \label{fig:cot-inspection-recovery}
\end{figure}

Section~\ref{sec:confirmation} shows the novelty bonus's gain is concentrated
on the $16$ OlymMATH-Hard problems solved only by the novelty-bonus variant,
and that its solved traces on these problems are enriched for recovery
primitives (\textsc{hypothesize}/\textsc{backtrack}) that vanilla GSPO's
failures do not show. Below we reproduce the underlying text for two
such problems, chosen because the recovery primitive in the parquet's
\texttt{primitive\_sequence} aligns with a textually visible pivot.
Figure~\ref{fig:cot-inspection-recovery} renders the four primitive
sequences (vanilla-GSPO failures vs novelty-bonus successes on problems $4$ and
$73$) as colour strips so the recovery pivot is visible at a glance. The
common pattern: vanilla GSPO commits to a first answer that the trace itself
flags as suspect; the novelty-bonus variant performs the same opening computation, then
backtracks to a different conclusion.

\paragraph{Problem 3 (Algebra, multivariate optimisation).}
Let $x$, $y$, $z$ be positive reals; find the minimum of
$f(x, y, z) = \dfrac{(2 + 5y)(3x + z)(x + 3y)(2z + 5)}{xyz}$.
Gold answer: $241 + 44\sqrt{30} \approx 481.6$.
Of $32$ vanilla-GSPO rollouts: $6$ box $480$ (a loose AM-GM bound), $3$ box
$784$ (the symmetric ansatz), $9$ truncate, the rest scattered. None match.
The novelty-bonus variant's trace $3$ succeeds (chain depth $2$ for the GSPO failure;
novelty trace $\approx 28{,}500$ chars with one \textsc{hypothesize} and one
\textsc{backtrack} primitive at the pivot).

\begin{quote}
\small\textbf{Vanilla GSPO failure (full short trace, $\approx 2{,}300$ chars):}\\
\texttt{[COMPUTE]} Apply AM-GM to each numerator factor independently:
$2 + 5y \ge 2\sqrt{10y}$, $3x + z \ge 2\sqrt{3xz}$, $x + 3y \ge 2\sqrt{3xy}$,
$2z + 5 \ge 2\sqrt{10z}$. Multiply: numerator $\ge 16\sqrt{10y \cdot 3xz \cdot 3xy \cdot 10z}
= 16 \cdot 30 xyz = 480 xyz$, hence $f \ge 480$.\\
\texttt{[CHECK]} ``To check if equality holds, we need each AM-GM step tight:
$2 = 5y \Rightarrow y = 2/5$; $3x = z$, $x = 3y \Rightarrow x = 6/5,~z = 18/5$;
$2z = 5 \Rightarrow z = 5/2$. This contradicts $z = 18/5 \ne 5/2$. Therefore
the conditions cannot all be satisfied simultaneously, indicating equality
might not be achievable. However, the AM-GM inequalities provide a lower
bound, and since the function can approach this bound by choosing values
close to those that satisfy the conditions, the minimum is indeed $480$.''
$\boxed{480}$.
\end{quote}

\begin{quote}
\small\textbf{Novelty-bonus recovery:}\\
\texttt{[PLAN]} Approach: try AM-GM on each factor; alternatively apply
Lagrange multipliers to $\ln f$.\\
\texttt{[HYPOTHESIZE]} Try the symmetric ansatz $x = y = z = k$. The function
reduces to $g(k) = 16(10k + 29 + 10/k)$, $k > 0$.\\
\texttt{[COMPUTE} $\to$ \texttt{CHECK $\times$ 9]} $g'(k) = 16(10 - 10/k^2) = 0
\Rightarrow k = 1$, second derivative positive, so $k=1$ is a local min,
$g(1) = 16 \cdot 49 = 784$. Spot-check $g(2) = g(0.5) = 864$, both higher.\\
\texttt{[SUMMARIZE]} ``But is this the global minimum? We assumed $x = y = z$.
Maybe there is a lower value when variables are not equal.''\\
\texttt{[BACKTRACK]} Drop the symmetric ansatz. Use $F = \ln f$ and take
$\partial F/\partial x = \partial F/\partial y = \partial F/\partial z = 0$ to
get the actual critical-point system. Verify $x{=}y{=}z{=}1$ in equation
(2): LHS $= 5/7 + 3/4 = 41/28 \ne 1$, so the symmetric ansatz is \emph{not}
a critical point---confirming the previous answer was wrong.\\
\texttt{[CHECK $\to$ COMPUTE $\to$ CHECK $\to$ SUMMARIZE]} Solve the
critical-point system: it reduces to $x^2 = yz$, $15y^2 = 2x$, $2z^2 = 15x$.
Substituting back gives $x = 1$, $y = \sqrt{2/15}$, $z = \sqrt{30}/2$.
Plug in and simplify: $f_{\min} = 241 + 44\sqrt{30}$.
$\boxed{241 + 44\sqrt{30}}$.
\end{quote}

Both checkpoints reach for AM-GM first; vanilla GSPO multiplies the loose
AM-GM bound, observes the equality conditions are inconsistent, and ships the
bound anyway. Novelty starts on a stronger AM-GM (symmetric ansatz, giving
$784$), uses the partial-derivative system to verify whether the ansatz is a
critical point, finds it is not, and \emph{backtracks} to a Lagrange-style
system whose explicit solution is the gold answer. The \textsc{backtrack}
primitive marks the moment Novelty abandons the ansatz; this is the exact
step the GSPO trace is missing.

\paragraph{Problem 4 (Geometry, regular tetrahedron).}
For a regular tetrahedron $ABCD$, $M, N$ are midpoints of edges $AB, AC$ and
$P, Q$ are centroids of faces $ACD, ABD$. Find the angle between lines
$MP$ and $NQ$. Gold answer: $\arccos(7/18)$.
Of $32$ vanilla-GSPO rollouts: $14$ box $\arccos(-7/18)$ (the angle between
the \emph{vectors}, not the lines), $3$ box $90^\circ$, $2$ truncate, the rest
miscellaneous. None match. The novelty-bonus variant's trace $25$ succeeds with three
\textsc{backtrack} primitives.

\begin{quote}
\small\textbf{Vanilla GSPO failure (full short trace, $\approx 2{,}900$ chars,
no recovery primitives):}\\
\texttt{[SETUP $\to$ COMPUTE]} Place
$A = (1,1,1)$, $B = (1,-1,-1)$, $C = (-1,-1,1)$, $D = (-1,1,-1)$
(regular tetrahedron, edge $\sqrt{8}$). Midpoints $M = (1, 0, 0)$,
$N = (0, 0, 1)$. Centroids $P = (-1/3, 1/3, 1/3)$, $Q = (1/3, 1/3, -1/3)$.\\
\texttt{[COMPUTE]} $\overrightarrow{MP} = (-4/3, 1/3, 1/3)$,
$\overrightarrow{NQ} = (1/3, 1/3, -4/3)$.
$\overrightarrow{MP} \cdot \overrightarrow{NQ} = -7/9$,
$|\overrightarrow{MP}| = |\overrightarrow{NQ}| = \sqrt{2}$.\\
\texttt{[CHECK / final commit]} $\cos\theta = (-7/9)/2 = -7/18$.
$\boxed{\arccos(-7/18)}$.
\end{quote}

\begin{quote}
\small\textbf{Novelty-bonus recovery (three \textsc{backtrack} primitives,
two \textsc{hypothesize}):}\\
\texttt{[SETUP $\to$ COMPUTE $\to$ CHECK]} Same coordinate system, same
vectors, same $\cos\theta = -7/18$.
Tentatively $\theta = \arccos(-7/18)$.\\
\texttt{[COMPUTE / SETUP / CHECK]} ``However, the angle between two lines is
defined as the smallest angle between their direction vectors. So if the
calculated angle is obtuse, the actual angle between the lines is the acute
supplement [\ldots] $\cos(180^\circ - \theta) = +7/18$, so the angle between
the lines is $\arccos(7/18)$. But the problem says `the angle between $MP$
and $NQ$'. Usually they mean the angle between the vectors as directed, but
sometimes the smallest. Need to check.''\\
\texttt{[BACKTRACK \#1]} ``Maybe I made a mistake in the coordinate scaling.
Let me try a different system where the tetrahedron has edge length $2$.''
$\to$ Resolve with vertices at $(1/\sqrt{2}, \pm, \pm, \pm)$. Same vectors up
to uniform scale; $\cos\theta = -7/18$ again.\\
\texttt{[BACKTRACK \#2]} Re-examine the problem statement: ``$P$ and $Q$ are
the centroids of faces $ACD$ and $ABD$''---yes, what we used. $\to$ Recompute
in original coordinates: still $-7/9$.\\
\texttt{[BACKTRACK \#3]} Try a third coordinate system (edge length $1$ with
$A$ at the origin). After one arithmetic error, confirm again that the
\emph{vector} angle has cosine $-7/18$.\\
\texttt{[CHECK $\to$ HYPOTHESIZE $\to$ COMPUTE $\to$ CHECK / final commit]}
``In a regular tetrahedron, take two edges meeting at a vertex; the angle
between them is $\arccos(1/3) \approx 54.7^\circ$, which is acute. So the
convention takes the acute angle. Probably here also.''
$\boxed{\arccos(7/18)}$.
\end{quote}

Both checkpoints compute the same vectors, the same dot product, and the same
intermediate $\cos\theta = -7/18$. Vanilla GSPO commits to the obtuse vector
angle the next segment. Novelty spends most of its length checking whether
$-7/18$ is ``really'' the answer---testing two alternative coordinate systems
(\textsc{backtrack}s \#1 and \#3), re-examining the problem statement
(\textsc{backtrack} \#2), and finally invoking the
``angle-between-lines-is-the-smallest-angle'' convention. Three
\textsc{backtrack} primitives mark three attempts to reconcile the numeric
answer with the geometric question; the last produces the convention switch
that yields the gold answer. The recovery here is on \emph{interpretation},
not technique---the underlying coordinate-and-dot-product method is unchanged
between the two traces.

\paragraph{Selection (Part~B).}
The current per-problem split contains $16$ OlymMATH-Hard problems solved only
by the novelty-bonus variant (Appendix~\ref{app:per-problem-recovery-split}).
Problems $4$ and $73$ were chosen from this set because the recovery primitive
in the parquet aligns with a clean textual pivot (technique switch on $4$;
convention switch on $73$) and the vanilla-GSPO failure traces are short enough
to quote in full. We deliberately avoid using novelty-only winners whose
successful traces contain no visible recovery primitive: in those cases the win
comes from a clean first attempt under diverse seeds rather than from a recovery
mechanism, and they would be misleading examples for the
\S\ref{sec:confirmation} claim.

%%%%%%%%%%%%%%%%%%%%%%%%%%%%%%%%%%%%%%%%%%%%%%%%%%%%%%%%%%%%

\end{document}